\documentclass[10pt,journal,compsoc]{IEEEtran}
\ifCLASSOPTIONcompsoc
  \usepackage[nocompress]{cite}
\else
  \usepackage{cite}
\fi
\ifCLASSINFOpdf
\else
\fi
\usepackage{amsmath, amssymb}
\usepackage{array}
\ifCLASSOPTIONcompsoc
 \usepackage[caption=false,font=footnotesize,labelfont=sf,textfont=sf]{subfig}
\else
 \usepackage[caption=false,font=footnotesize]{subfig}
\fi
\usepackage{fixltx2e}
\usepackage{dblfloatfix}
\ifCLASSOPTIONcaptionsoff
 \usepackage[nomarkers]{endfloat}
\let\MYoriglatexcaption\caption
\renewcommand{\caption}[2][\relax]{\MYoriglatexcaption[#2]{#2}}
\fi
\usepackage{url}
\usepackage{graphicx}
\usepackage{bbm}
\usepackage{multirow}
\usepackage[noend]{algpseudocode}
\usepackage{algorithmicx,algorithm}
\usepackage[misc]{ifsym}
\usepackage{color}
\usepackage[table]{xcolor}
\usepackage{booktabs}

\begin{document}
%
% paper title
% Titles are generally capitalized except for words such as a, an, and, as,
% at, but, by, for, in, nor, of, on, or, the, to and up, which are usually
% not capitalized unless they are the first or last word of the title.
% Linebreaks \\ can be used within to get better formatting as desired.
% Do not put math or special symbols in the title.
\title{Mutual Information Guided Optimal Transport for Unsupervised Visible-Infrared Person Re-identification}
% Rethinking Unsupervised Visible-Infrared Person Re-Identification Through Mutual Information Theory
%
%
% author names and IEEE memberships
% note positions of commas and nonbreaking spaces ( ~ ) LaTeX will not break
% a structure at a ~ so this keeps an author's name from being broken across
% two lines.
% use \thanks{} to gain access to the first footnote area
% a separate \thanks must be used for each paragraph as LaTeX2e's \thanks
% was not built to handle multiple paragraphs
%
%
%\IEEEcompsocitemizethanks is a special \thanks that produces the bulleted
% lists the Computer Society journals use for "first footnote" author
% affiliations. Use \IEEEcompsocthanksitem which works much like \item
% for each affiliation group. When not in compsoc mode,
% \IEEEcompsocitemizethanks becomes like \thanks and
% \IEEEcompsocthanksitem becomes a line break with idention. This
% facilitates dual compilation, although admittedly the differences in the
% desired content of \author between the different types of papers makes a
% one-size-fits-all approach a daunting prospect. For instance, compsoc 
% journal papers have the author affiliations above the "Manuscript
% received ..."  text while in non-compsoc journals this is reversed. Sigh.

%\author{Michael~Shell,~\IEEEmembership{Member,~IEEE,}
 %       John~Doe,~\IEEEmembership{Fellow,~OSA,}
  %      and~Jane~Doe,~\IEEEmembership{Life~Fellow,~IEEE}% <-this % stops a space
\author{Zhizhong Zhang$^*$,  Jiangming Wang$^*$, Xin Tan, Yanyun Qu,~\IEEEmembership{Member,~IEEE}, Junping Wang, \\ Yong Xie, Yuan Xie$^\dagger$,~\IEEEmembership{Member,~IEEE}

\IEEEcompsocitemizethanks{
%\IEEEcompsocthanksitem M. Shell was with the Departmentof Electrical and Computer Engineering, Georgia Institute of Technology, Atlanta,GA, 30332.\protect\\
% note need leading \protect in front of \\ to get a newline within \thanks as
% \\ is fragile and will error, could use \hfil\break instead.
%E-mail: see http://www.michaelshell.org/contact.html
%\IEEEcompsocthanksitem J. Doe and J. Doe are with Anonymous University.% <-this % stops an unwanted space
%\thanks{Manuscript received April 19, 2005; revised August 26, 2015.}
\IEEEcompsocthanksitem $^*$: Equal contribution. $^\dagger$: Corresponding Author.
\IEEEcompsocthanksitem Z. Zhang, J. Wang, X. Tan and Y. Xie are with the School of Computer Science and
Technology, East China Normal University, Shanghai, China. E-mail: \{zzzhang, yxie, xtan\}@cs.ecnu.edu.cn; 51215901073@stu.ecnu.edu.cn.

\IEEEcompsocthanksitem Y. Qu is with the School of Informatics, Xiamen University, Xiamen, China. E-mail: yyqu@xmu.edu.cn.

\IEEEcompsocthanksitem J. Wang is with the Institute of Automation Chinese Academy of Sciences, Beijing, China. E-mail: junping.wang@ia.ac.cn.

\IEEEcompsocthanksitem Y. Xie is with Nanjing University of Posts and Telecommunications, Nanjing, China. E-mail: yongxie@njupt.edu.cn.
}
}

% note the % following the last \IEEEmembership and also \thanks - 
% these prevent an unwanted space from occurring between the last author name
% and the end of the author line. i.e., if you had this:
% 
% \author{....lastname \thanks{...} \thanks{...} }
%                     ^------------^------------^----Do not want these spaces!
%
% a space would be appended to the last name and could cause every name on that
% line to be shifted left slightly. This is one of those "LaTeX things". For
% instance, "\textbf{A} \textbf{B}" will typeset as "A B" not "AB". To get
% "AB" then you have to do: "\textbf{A}\textbf{B}"
% \thanks is no different in this regard, so shield the last } of each \thanks
% that ends a line with a % and do not let a space in before the next \thanks.
% Spaces after \IEEEmembership other than the last one are OK (and needed) as
% you are supposed to have spaces between the names. For what it is worth,
% this is a minor point as most people would not even notice if the said evil
% space somehow managed to creep in.

% The paper headers
\markboth{}%
{Shell \MakeLowercase{\textit{et al.}}: Bare Demo of IEEEtran.cls for Computer Society Journals}
% The only time the second header will appear is for the odd numbered pages
% after the title page when using the twoside option.
% 
% *** Note that you probably will NOT want to include the author's ***
% *** name in the headers of peer review papers.                   ***
% You can use \ifCLASSOPTIONpeerreview for conditional compilation here if
% you desire.

% The publisher's ID mark at the bottom of the page is less important with
% Computer Society journal papers as those publications place the marks
% outside of the main text columns and, therefore, unlike regular IEEE
% journals, the available text space is not reduced by their presence.
% If you want to put a publisher's ID mark on the page you can do it like
% this:
%\IEEEpubid{0000--0000/00\$00.00~\copyright~2015 IEEE}
% or like this to get the Computer Society new two part style.
%\IEEEpubid{\makebox[\columnwidth]{\hfill 0000--0000/00/\$00.00~\copyright~2015 IEEE}%
%\hspace{\columnsep}\makebox[\columnwidth]{Published by the IEEE Computer Society\hfill}}
% Remember, if you use this you must call \IEEEpubidadjcol in the second
% column for its text to clear the IEEEpubid mark (Computer Society jorunal
% papers don't need this extra clearance.)

% use for special paper notices
%\IEEEspecialpapernotice{(Invited Paper)}

% for Computer Society papers, we must declare the abstract and index terms
% PRIOR to the title within the \IEEEtitleabstractindextext IEEEtran
% command as these need to go into the title area created by \maketitle.
% As a general rule, do not put math, special symbols or citations
% in the abstract or keywords.
\IEEEtitleabstractindextext{%
\begin{abstract}
Unsupervised visible infrared person re-identification (USVI-ReID) is a challenging retrieval task that aims to retrieve cross-modality pedestrian images without using any label information. In this task, the large cross-modality variance makes it difficult to generate reliable cross-modality labels, and the lack of annotations also provides additional difficulties for learning modality-invariant features. In this paper, we first deduce an optimization objective for unsupervised VI-ReID based on the mutual information between the model's cross-modality input and output. With equivalent derivation, three learning principles, \textit{i.e.}, "Sharpness" (entropy minimization), "Fairness" (uniform label distribution), and "Fitness" (reliable cross-modality matching) are obtained. Under their guidance, we design a loop iterative training strategy alternating between model training and cross-modality matching. In the matching stage, a uniform prior guided optimal transport assignment ("Fitness", "Fairness") is proposed to select matched visible and infrared prototypes. In the training stage, we utilize this matching information to introduce prototype-based contrastive learning for minimizing the intra- and cross-modality entropy ("Sharpness"). Extensive experimental results on benchmarks demonstrate the effectiveness of our method, \textit{e.g.}, 60.6\% and 90.3\% of Rank-1 accuracy on SYSU-MM01 and RegDB without any annotations. 
\end{abstract}

% Note that keywords are not normally used for peerreview papers.
\begin{IEEEkeywords}
Unsupervised Learning, Visible Infrared Person Re-Identification, Optimal Transport, Mutual Information.
\end{IEEEkeywords}}

% make the title area
\maketitle

% To allow for easy dual compilation without having to reenter the
% abstract/keywords data, the \IEEEtitleabstractindextext text will
% not be used in maketitle, but will appear (i.e., to be "transported")
% here as \IEEEdisplaynontitleabstractindextext when the compsoc 
% or transmag modes are not selected <OR> if conference mode is selected 
% - because all conference papers position the abstract like regular
% papers do.
\IEEEdisplaynontitleabstractindextext
% \IEEEdisplaynontitleabstractindextext has no effect when using
% compsoc or transmag under a non-conference mode.

% For peer review papers, you can put extra information on the cover
% page as needed:
% \ifCLASSOPTIONpeerreview
% \begin{center} \bfseries EDICS Category: 3-BBND \end{center}
% \fi
%
% For peerreview papers, this IEEEtran command inserts a page break and
% creates the second title. It will be ignored for other modes.
\IEEEpeerreviewmaketitle

\IEEEraisesectionheading{\section{Introduction}\label{sec:introduction}} 
The purpose of person re-identification (ReID) is to match the pedestrian images across multiple non-overlapping camera views. Most ReID methods \cite{AGW,Nformer,LPN,SECRET,ISE,PPLR,STS} focus on the daytime setting due to the easy acquisition of colorful visible images. However, with the development of sensor technology, \textit{e.g.}, infrared cameras, clear images can also be captured under poor illumination conditions. Therefore, visible-infrared person re-identification (VI-ReID) emerges as a critical enabler for night intelligent monitoring systems, which allows us to retrieve the target images (visible or infrared) when given a query image from another modality. Due to the significant difference in sensing processes, visible-infrared heterogeneous images have very large appearance and color variations. Hence, it will be much more challenging than the conventional visible person re-identification task.

\begin{figure}[!t]
	\centering
	\includegraphics[scale=0.31]{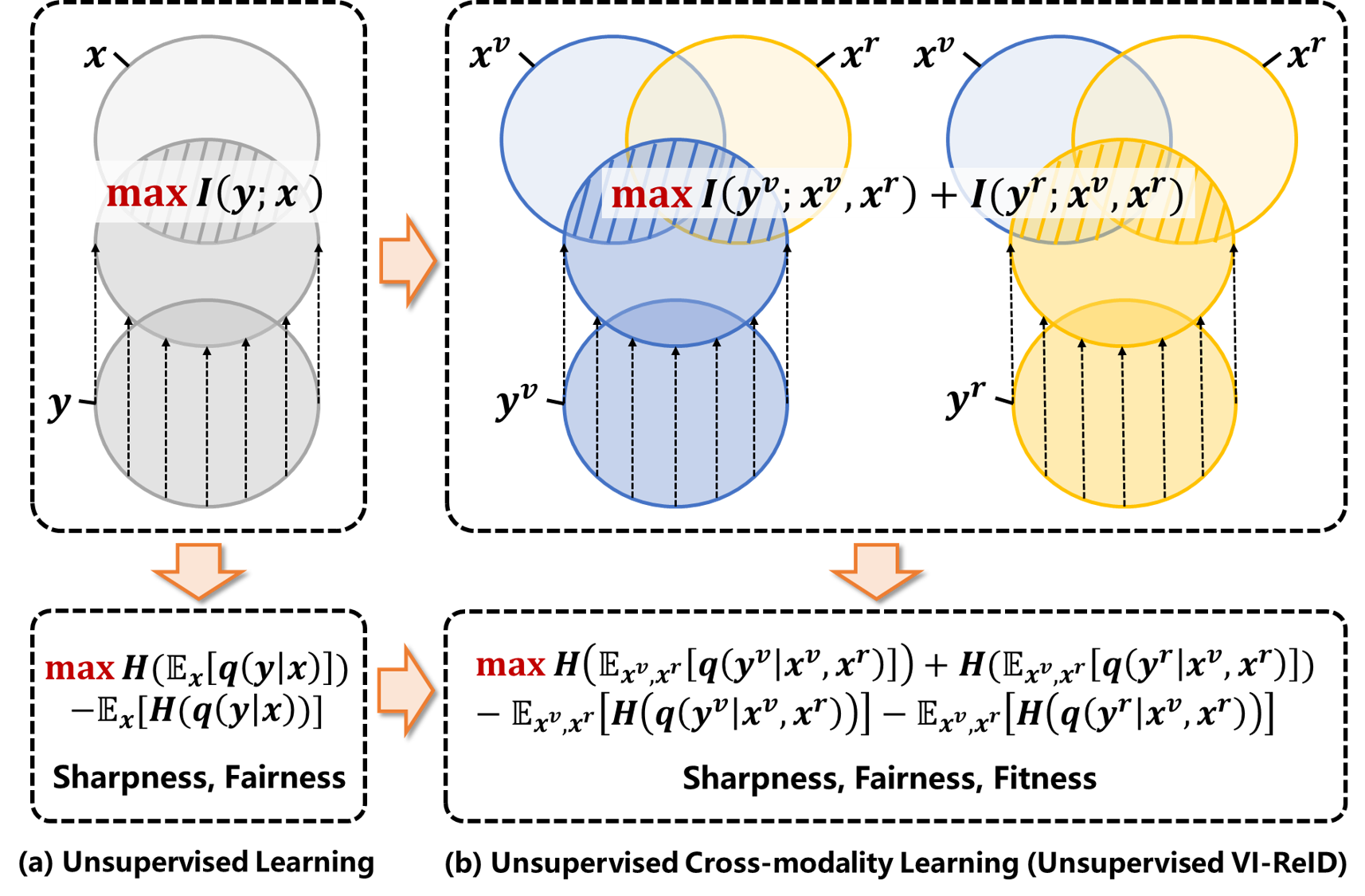}
	\caption{(a) One way to realize the unsupervised learning is to maximize the mutual information between the model's input $\boldsymbol{x}$ and output $\boldsymbol{y}$ \cite{ReMixMatch}, (\textit{i.e.}, "Sharpness" and "Fairness"). (b) For unsupervised VI-ReID task, we additionally mine the modality-consistent mutual information (\textit{i.e.}, maximize the mutual information between the matched cross-modality input pairs $\boldsymbol{x}^{v},\boldsymbol{x}^{r}$ and the corresponding outputs $\boldsymbol{y}^{v},\boldsymbol{y}^{r}$, where $v$ and $r$ denote visible and infrared modality respectively) (\textit{i.e.}, "Fitness").}
	\label{Fig:introduction}
\end{figure}

Recently, impressive progress \cite{DDAG,AGW,Hi-CMD,LbA,NFS,MPANet,CAJL,FMCNet,CMT,MSCLNet} in VI-ReID has been made to reduce the cross-modality discrepancy. A common practice is to align the visible and infrared images on both image and feature level \cite{CMT}, either through converting images from one modality to another \cite{JSIA-ReID,Hi-CMD}, or using cross-modality metric \cite{AGW,CAJL,MPANet} and architecture \cite{DDAG,MSCLNet}. However, one of the important ingredients to their success is the availability of well-annotated training sets. These training sets need extensive labeling efforts, especially for the infrared subsets, as it is difficult for annotators to recognize the identities without color information. 

To deal with this issue, a series of works \cite{H2H,OTLA-ReID,ADCA,CHCR,DOTLA,MBCCM,CCLNet,PGM} have already tried to acquire a discriminative VI-ReID model without infrared label information or complete annotation. However, existing works often lack clear and explainable learning objectives, where the significant cross-modality discrepancy leads to difficulties in generating reliable cross-modality pseudo labels and the lack of annotations also provides additional difficulties in learning discriminative modality-invariant features. Here are two critical issues in this area: 1) how to assign reliable cross-modality pseudo labels? 2) how to learn such modality-invariant features?
 
In this paper, we provide a learnable and explainable objective for the unsupervised VI-ReID task based on mutual information theory. As shown in Fig. \ref{Fig:introduction}, this objective is deduced from the ReMixMatch \cite{ReMixMatch}. We further extend it to cross-modal unsupervised representation, which additionally mines the modality-consistent mutual information. According to it, we obtain three learning principles (\textit{i.e.}, "Sharpness", "Fairness" and "Fitness") to guide the optimization direction of the unsupervised VI-ReID task without any annotations or extra labeled visible datasets. In this framework,  "Fitness" refers to  matching the cross-modality samples, "Fairness" refers to generating uniform pseudo label distribution, and "Sharpness" refers to learning identity-aware features.

%In detail, to assign the reliable cross-modality pseudo labels, our approach enables to find the cross-modality correspondence, {\it i.e., } the matching relationship between each visible prototype and infrared prototype, under the guidance of ``Sharpness", ``Fairness" and ``Fitness".  Note that this is different from the common practice in unsupervised ReID, where each instance is transported to a cluster center. As a result,  it brings a new issue that the number of visible prototypes is different from the infrared ones leading to a \textbf{one-to-many matching}, although its formulation seems similar.  We propose an opposite transport strategy by exchanging suppliers and demanders and therefore establish robust cross-modality correspondence. 

In our implementation, we design a loop iterative training strategy alternating between model training and cross-modality matching. At the beginning of each training epoch, DBSCAN \cite{DBSCAN} is used to generate coarse pseudo labels for each modality data. Then we obtain a cluster-aware prototype by averaging the features belonging to the same pseudo label. To reach "Fitness", we formulate this cross-modality prototype matching problem as a dual optimal transport problem, such that the cross-modality prototypes with higher transport probability would be merged together, {\it i.e.,} cross-modality correspondence. To overcome all prototypes being grouped into one, we introduce "Fairness" by adding the uniform prior to this optimal transport process. With this correspondence information, we design various prototype-based contrastive learning losses as an equivalent substitute for entropy minimization, allowing us to reach "Sharpness". %Finally, we propose a revised prediction alignment learning loss inspired by \cite{OTLA-ReID} to eliminate the negative effect brought by inaccurately matched cross-modality data. 
Alternating between cross-modality matching and learning, our model can gradually self-learn useful information to produce discriminative feature representation for unlabeled cross-modal data. Our contributions can be summarized as follows:

$\bullet$ Based on the analysis of mutual information theory, we deduce the theoretical objective for the unsupervised VI-ReID task. It can be further transferred into three optimizable principles (\textit{i.e.}, "Sharpness", "Fairness" and "Fitness"). 

$\bullet$ We design a loop iterative training strategy alternating between model training and cross-modality matching, that simultaneously allows for generating reliable cross-modality pseudo labels and learning modality-invariant features.

%Three key components are proposed: (1) Optimal Transport Prototype Assignment, which enables us to gradually match the unlabeled cross-modality data. (2) Prototype-Based Contrastive Learning, which is an equivalent substitute for entropy minimization. (3) Cross Prediction Alignment Learning which can reduce the negative effects brought by inaccurately matched cross-modality data.

$\bullet$ We conduct extensive experiments on widely adopted VI-ReID benchmarks. Empirical results show that our proposed method achieves great improvement against a series of state-of-the-art unsupervised learning and unsupervised domain adaption ReID (VI-ReID) methods, also highly comparable against fully supervised VI-ReID methods.

\section{Related Work}
\subsection{Person Re-Identification}
\textbf{Unsupervised Learning Person ReID (USL-ReID).} Unsupervised person re-identification is a single-modality ReID task, which aims to train a model with only unlabeled visible pedestrian data. In this setting, nearly all of the methods \cite{SpCL,MMCL,MetaCam,BUC,HCD,ISE,MCL,PPLR,STS,DCMIP} are pseudo labels based, which then establish a bridge towards the supervised manner via these pseudo labels. For example, several works \cite{SpCL,MetaCam,ISE,PPLR,STS,DCMIP} try to obtain pseudo labels with the traditional clustering method(\textit{e.g.}, DBSCAN or K-means) and then gradually improve the quality of pseudo labels by refining clusters as the training goes on. Specifically, ISE \cite{ISE} uses the progressive linear interpolation of KNN samples to update the clusters. While \cite{SpCL,STS,DCMIP} gets the help of contrastive learning loss with the special design of clustering memory bank. As for MetaCam \cite{MetaCam}, it benefits from camera-aware meta-learning to produce discriminative clustering centroids. Besides, some hierarchical clustering ways \cite{BUC,HCD} with merge-split cluster's refinement are proposed to obtain high-quality pseudo labels from another perspective.

\noindent\textbf{Supervised Visible-Infrared Person ReID (SVI-ReID).} It is a challenging cross-modality image retrieval problem, which aims to match the pedestrian images captured by visible and infrared cameras. Up to now, impressive progress has been made in it. To this end, most works \cite{DDAG,cm-SSFT,MPANet,LbA,FMCNet,CMT,MSCLNet,DEEN,SEFL,PartMix,MUN} in this community try to enhance the feature discrimination by introducing novel network architectures (\textit{e.g.}, graph convolution network, non-local module, transformer or self-designed modality fusion model). Some of these works \cite{MPANet,LbA,MSCLNet,SEFL,PartMix} also can discover nuanced but discriminative clues (\textit{e.g.}, part or patch-level information) for both modalities so as to improve the quality of feature representation. While, another line of works \cite{D2RL,XModality,AlignGAN,JSIA-ReID,Hi-CMD,MID} attempts to excavate modality-invariant information via cross-modality image generation or mix-up manipulation. Besides, \cite{cmGAN,BDTR,PSE,LbA,MAUM} have optimized metric learning losses (\textit{e.g.}, triplet loss or memory-based contrastive learning loss) adapting to cross-modality learning. 

\noindent\textbf{Unsupervised Visible-Infrared Person ReID (USVI-ReID)}. Recently, this task has gradually drawn great attention due to the laboring efforts of annotation, especially for infrared data without color information. Some works \cite{H2H,OTLA-ReID} try to solve this problem with the assistance of other labeled visible datasets (\textit{e.g.}, Market-1501 \cite{Market-1501}). Specifically, OTLA-ReID \cite{OTLA-ReID} utilizes the standard unsupervised domain adaptation technique to generate the pseudo labels for the visible subset with the help of well-annotated visible datasets, and then leverage optimal transport theory to facilitate cross-modality learning. Other works \cite{ADCA,CHCR,DOTLA,MBCCM,PGM} directly apply self-supervised learning on unlabeled cross-modality datasets. Among them, MBCCM \cite{MBCCM} and PGM \cite{PGM} employ bipartite graph matching to uncover cross-modality correspondence. Besides, CCLNet \cite{CCLNet} incorporates both image and text information of pedestrians into the unsupervised cross-modality model training by using the pre-trained vision-language model (\textit{i.e.}, CLIP \cite{CLIP}).

% \subsection{Unsupervised Domain Adaptation Person ReID}
% The goal of unsupervised domain adaptation \cite{TransferLearningSurvey} is to transfer the knowledge from a labeled source domain to an unlabeled target domain. The recent application in ReID community (UDA-ReID) can be regarded as an open set task, where label spaces between two domains do not overlap. UDA-ReID can be roughly divided into three categories. The first category \cite{ECN,GPP,D-MMD} attempts to reduce the domain gap by digging up positive or negative pairs by using this open set prior. The second category \cite{SSG,MMT,SpCL,GLT,SECRET} has adopted unsupervised clustering methods to mine the supervision from the target domain. Most methods of this category \cite{SSG,MMT,GLT,SECRET} firstly pretrains a model on source domain data by supervised methods and then alternates between clustering and fine-tuning in the target domain. While SpCL \cite{SpCL} employs the self-paced clustering strategy without the source domain pretraining stage. The third category \cite{SPGAN,PTGAN,HHL} learns the domain invariant information by generating cross-domain images with the style transfer technique.

\subsection{Optimal Transport} 
Optimal transport (OT) is often used to find connections between variables or measure the probability distribution distance, which already has obtained increasing attention in machine learning. One of the widely used implementations is the Sinkhorn-Knopp algorithm \cite{Sinkhorn-Knopp}, which decreases the computation cost of OT problems significantly. SeLa \cite{OT} has extended OT to self-supervised learning. In this framework, they alternate between the following two steps: 1) Making use of the Sinkhorn-Knopp algorithm to produce pseudo labels for unlabeled data. 2) Doing classification with current pseudo labels. In fact, SeLa \cite{OT} is a clustering-based self-supervised approach, which aims to find a good pretraining model. The most related work to ours is OTLA-ReID \cite{OTLA-ReID}. However, it mainly focuses on the semi-supervised task and applies OT for a sample-label assignment problem with the help of the classifier. Obviously, it's much harder to define the proper classifier without any real label information. Thus it doesn't work well for the unsupervised task. By contrast, our proposed method mainly focuses on the unsupervised task and applies OT for a cross-modality prototype assignment problem without using a classifier, which also achieves better performance than OTLA-ReID.

% It can serve as theoretical guidance to ensure that the learned feature representation contains relevant information for the downstream task. 
% However, estimating mutual information is a challenging problem, especially for high-dimensional feature representation extracted by deep neural networks.

\section{Preliminary: Mutual Information}
In this section, we first provide a brief review of the mutual information theory \cite{IBM} in the context of unsupervised and semi-supervised learning. %Then we extend it to unsupervised cross-modal representation learning.} %Then we would like to propose a learnable objective based on mutual information for the unsupervised VI-ReID task. This objective function can be decomposed into three learning principles, which guide the optimization direction.} 
Given input data $\boldsymbol{x}$, some works \cite{CPC,InfoMax,MIM,MIB} start from the information bottleneck principle with objective $\max I(\boldsymbol{x};\boldsymbol{z})$. This objective function can be seen as a process of feature compression, which makes the representation $\boldsymbol{z}$ contain as much useful information from the input data $\boldsymbol{x}$ as possible. However, ReMixMatch \cite{ReMixMatch} and related works \cite{UMI,ReMixMatch,MIRS,RDA} directly maximize the mutual information between the model's input $\boldsymbol{x}$ and output $\boldsymbol{y}$ (\textit{i.e.}, predicted label in unsupervised learning):
\begin{equation}
\begin{aligned}
    \max\ I(\boldsymbol{y};\boldsymbol{x}) = \iint p(\boldsymbol{y},\boldsymbol{x}){\rm log}(\frac{p(\boldsymbol{y},\boldsymbol{x})}{p(\boldsymbol{y})p(\boldsymbol{x})}) d\boldsymbol{y}d\boldsymbol{x}.
    \label{mi_remixmatch}
\end{aligned}
\end{equation}

According to variational inference, we can obtain the variational lower bound of $I(\boldsymbol{x};\boldsymbol{y})$ with a form of the negative entropy term plus regularization term:
\begin{equation}
\begin{aligned}
    I(\boldsymbol{y};\boldsymbol{x}) \ge \underbrace{H(\mathbbm{E}_{\boldsymbol{x}}[q(\boldsymbol{y}|\boldsymbol{x})])}_{\rm Regularization}-\underbrace{\mathbbm{E}_{\boldsymbol{x}}[H(q(\boldsymbol{y}|\boldsymbol{x}))]}_{\rm Entropy} \overset{{\rm def}}{=} \tilde{I}(\boldsymbol{y};\boldsymbol{x}),
    \label{mi_remixmatch_reduction}
\end{aligned}
\end{equation}
where $q(\cdot)$ is an arbitrary variational distribution, $H(\cdot)$ denotes the entropy, and $\mathbbm{E}(\cdot)$ denotes the expectation value. Note that Eq. (\ref{mi_remixmatch_reduction}) is actually a refined objective deduced from ReMixMatch (\textit{i.e.}, we use variational distribution $q(\boldsymbol{y}|\boldsymbol{x})$ to replace $p(\boldsymbol{y}|\boldsymbol{x})$ to get a more rigorous form). The $q(\cdot)$ here can be explained as the class probability vector, \textit{e.g.}, classifier's prediction.

\section{Methodology}
\subsection{Problem Formulation and Overview}
Suppose we are given a collection of cross-modality pedestrian images  $\mathcal{X}=\{\mathcal{V},\mathcal{R}\}$, where $\mathcal{V}$ = $\left\{\boldsymbol{x}_{k}^{v}\right \}_{k=1}^{N^{v}}$ and $\mathcal{R}$ = $\left\{\boldsymbol{x}_{k}^{r}\right \}_{k=1}^{N^{r}}$ denote the visible and infrared images with $N^{v}$ and $N^{r}$ samples, respectively. 
%Due to the characteristics of the 
In our unsupervised task, we do not have any real label information. 

\textbf{Mutual Information Guided Learning Objective.} We modify the learning objective of Eq. (\ref{mi_remixmatch}) to extend it for unsupervised cross-modal representation:
\begin{equation}
\begin{aligned}
    \max\ I(\boldsymbol{y}^{v};\boldsymbol{x}^{v}) + I(\boldsymbol{y}^{v};\boldsymbol{x}^{r}) + I(\boldsymbol{y}^{r};\boldsymbol{x}^{r}) + I(\boldsymbol{y}^{r};\boldsymbol{x}^{v}),
    \label{objective1}
\end{aligned}
\end{equation}
where $\boldsymbol{x}^{v},\boldsymbol{x}^{r}$ denote the hypothetically matched cross-modality input pair, and $\boldsymbol{y}^{v},\boldsymbol{y}^{r}$ denote corresponding model's predicted label, \textit{e.g.}, pseudo label. Eq. (\ref{objective1}) implies that the model's output should contain the essential information for both intra- and cross-modality input of the same identity.

We can change the Eq. (\ref{objective1}) into a both simple and clear form by the following theorem:

\noindent\textbf{Theorem 1.} \textit{For random variables $\boldsymbol{x}^{1},\boldsymbol{x}^{2},\boldsymbol{y}^{1},\boldsymbol{y}^{2}$, we can observe the following correlation:
\begin{equation}
\begin{aligned}
    &\max\ I(\boldsymbol{x}^{1};\boldsymbol{y}^{1}) + I(\boldsymbol{x}^{2};\boldsymbol{y}^{1}) + I(\boldsymbol{x}^{1};\boldsymbol{y}^{2}) + I(\boldsymbol{x}^{2};\boldsymbol{y}^{2}) \\ &\Leftrightarrow \max\ I(\boldsymbol{y}^{1};\boldsymbol{x}^{1},\boldsymbol{x}^{2}) + I(\boldsymbol{y}^{2};\boldsymbol{x}^{1},\boldsymbol{x}^{2}),
\end{aligned}
\end{equation}
where $\boldsymbol{y}^{1},\boldsymbol{y}^2$ are dependent on the  $\boldsymbol{x}^{1},\boldsymbol{x}^{2}$, respectively.}

According to Theorem 1, we can obtain the following equivalent equation (see Appendix A.1 for detailed proof):
\begin{equation}
\begin{aligned}
    \max\ I(\boldsymbol{y}^{v};\boldsymbol{x}^{v},\boldsymbol{x}^{r}) + I(\boldsymbol{y}^{r};\boldsymbol{x}^{v},\boldsymbol{x}^{r}),
    \label{objective2}
\end{aligned}
\end{equation}
where $I(\boldsymbol{x}^{v},\boldsymbol{x}^{r};\boldsymbol{y}^{v})$ or $I(\boldsymbol{x}^{v},\boldsymbol{x}^{r};\boldsymbol{y}^{r})$ represents that the information contained in $\boldsymbol{y}^{v}$ or $\boldsymbol{y}^{r}$ and shared with the union of $\boldsymbol{x}^{v}$ and $\boldsymbol{x}^{r}$. That indicates the learning objective additionally mines the modality-consistent mutual information compared with pure unsupervised learning, and hence Eq. (\ref{objective2}) achieves the consistent goal of unsupervised and cross-modality learning.

To facilitate optimization of Eq. (\ref{objective2}) through model training, we proposed the following theorem:

\noindent\textbf{Theorem 2.} \textit{For random variables $\boldsymbol{x},\boldsymbol{y},\boldsymbol{z}$, if $\boldsymbol{z}$ is dependent on the union of $\boldsymbol{x}$ and $\boldsymbol{y}$, \textit{i.e.}, $p(\boldsymbol{z}|\boldsymbol{x},\boldsymbol{y}) > 0$, then we can obtain the variational lower bound $\tilde{I}(\boldsymbol{z};\boldsymbol{x},\boldsymbol{y})$ of $I(\boldsymbol{z};\boldsymbol{x},\boldsymbol{y})$ with KL divergence constraint ${\rm KL}(p(\boldsymbol{z}|\boldsymbol{x},\boldsymbol{y}) || q(\boldsymbol{z}|\boldsymbol{x},\boldsymbol{y})) = \int p(\boldsymbol{z}|\boldsymbol{x},\boldsymbol{y}){\rm log}(\frac{p(\boldsymbol{z}|\boldsymbol{x},\boldsymbol{y})}{q(\boldsymbol{z}|\boldsymbol{x},\boldsymbol{y})}) d\boldsymbol{z} \ge 0$:
\begin{equation}
\begin{aligned}
    I(\boldsymbol{z};\boldsymbol{x},\boldsymbol{y}) &\ge H(\mathbbm{E}_{\boldsymbol{x},\boldsymbol{y}}[q(\boldsymbol{z}|\boldsymbol{x},\boldsymbol{y})]) - \mathbbm{E}_{\boldsymbol{x},\boldsymbol{y}}[H(q(\boldsymbol{z}|\boldsymbol{x},\boldsymbol{y}))] \\ &\overset{{\rm def}}{=} \tilde{I}(\boldsymbol{z};\boldsymbol{x},\boldsymbol{y}),
\end{aligned}
\end{equation}
where $q(\boldsymbol{z}|\boldsymbol{x},\boldsymbol{y})$ is an arbitrary variational distribution.}

With the help of Theorem 2, we can derive the following corresponding variational lower bounds of $I(\boldsymbol{y}^{v};\boldsymbol{x}^{v},\boldsymbol{x}^{r})$ and $I(\boldsymbol{y}^{r};\boldsymbol{x}^{v},\boldsymbol{x}^{r})$, \textit{i.e.}, $ \tilde{I}(\boldsymbol{y}^{v};\boldsymbol{x}^{v},\boldsymbol{x}^{r})$ and $\tilde{I}(\boldsymbol{y}^{r};\boldsymbol{x}^{v},\boldsymbol{x}^{r})$ (see Appendix A.2 for detailed proof):
\begin{equation}
\begin{aligned}
    I(\boldsymbol{y}^{v};\boldsymbol{x}^{v},&\boldsymbol{x}^{r}) \ge H(\mathbbm{E}_{\boldsymbol{x}^{v},\boldsymbol{x}^{r}}[q(\boldsymbol{y}^{v}|\boldsymbol{x}^{v},\boldsymbol{x}^{r})]) - \\ &\mathbbm{E}_{\boldsymbol{x}^{v},\boldsymbol{x}^{r}}[H(q(\boldsymbol{y}^{v}|\boldsymbol{x}^{v},\boldsymbol{x}^{r}))] \overset{{\rm def}}{=} \tilde{I}(\boldsymbol{y}^{v};\boldsymbol{x}^{v},\boldsymbol{x}^{r}),
\end{aligned}
\label{objective3}
\end{equation}
\begin{equation}
\begin{aligned}
    I(\boldsymbol{y}^{r};\boldsymbol{x}^{v},&\boldsymbol{x}^{r}) \ge H(\mathbbm{E}_{\boldsymbol{x}^{v},\boldsymbol{x}^{r}}[q(\boldsymbol{y}^{r}|\boldsymbol{x}^{v},\boldsymbol{x}^{r})]) - \\ &\mathbbm{E}_{\boldsymbol{x}^{v},\boldsymbol{x}^{r}}[H(q(\boldsymbol{y}^{r}|\boldsymbol{x}^{v},\boldsymbol{x}^{r}))] \overset{{\rm def}}{=} \tilde{I}(\boldsymbol{y}^{r};\boldsymbol{x}^{v},\boldsymbol{x}^{r}).
\end{aligned}
\label{objective4}
\end{equation}

Thus, optimizing Eq. (\ref{objective2}) is equivalent to optimizing the variational lower bounds:
\begin{equation}
\begin{aligned}
    \max\ &H(\mathbbm{E}_{\boldsymbol{x}^{v},\boldsymbol{x}^{r}}[q(\boldsymbol{y}^{v}|\boldsymbol{x}^{v},\boldsymbol{x}^{r})]) + H(\mathbbm{E}_{\boldsymbol{x}^{v},\boldsymbol{x}^{r}}[q(\boldsymbol{y}^{r}|\boldsymbol{x}^{v},\boldsymbol{x}^{r})]) \\ - &\mathbbm{E}_{\boldsymbol{x}^{v},\boldsymbol{x}^{r}}[H(q(\boldsymbol{y}^{v}|\boldsymbol{x}^{v},\boldsymbol{x}^{r}))] - \mathbbm{E}_{\boldsymbol{x}^{v},\boldsymbol{x}^{r}}[H(q(\boldsymbol{y}^{r}|\boldsymbol{x}^{v},\boldsymbol{x}^{r}))].
\label{objective5}
\end{aligned}
\end{equation}

Note that the first two terms of Eq. (\ref{objective5}) encourage an average label distribution across the entire training set, which we named as "Fairness". The third and fourth terms of Eq. (\ref{objective5}), {\it w.r.t} the minimization of entropy, suggests the model should be confident and discriminative in some samples, which we named as "Sharpness". Besides, because of the above mentioned hypothesis with $\boldsymbol{x}^{v}$ and $\boldsymbol{x}^{r}$, {\it i.e.,} cross-modality correspondence, the unlabeled cross-modality data should be matched during the training process, which we named as "Fitness". These three principles guide our proposed loop iterative training strategy.

\textbf{Estimation of $q(\cdot)$.} For the semi-supervised learning setting, such as ReMixMatch, $q(\cdot)$ can be transferred into the classifier's prediction with softmax manipulation. However, we cannot define a proper parameterized classifier under the unsupervised setting. Here we use clustering methods to obtain prototypes (\textit{e.g.}, mean of latent feature representations) to imitate the column parameter vectors of the classifier. Then we transfer $q(\cdot)$ into a prototype-based cosine similarity vector with softmax manipulation, inspired by few-shot learning\cite{PCFIC}. %This intuition comes from the few-shot image classification task, where the prototype classifier is frequently used and achieves comparable results with the parameterized classifier \cite{PCFIC}.

\begin{figure*}[!t]
	\centering
	\includegraphics[scale=0.52]{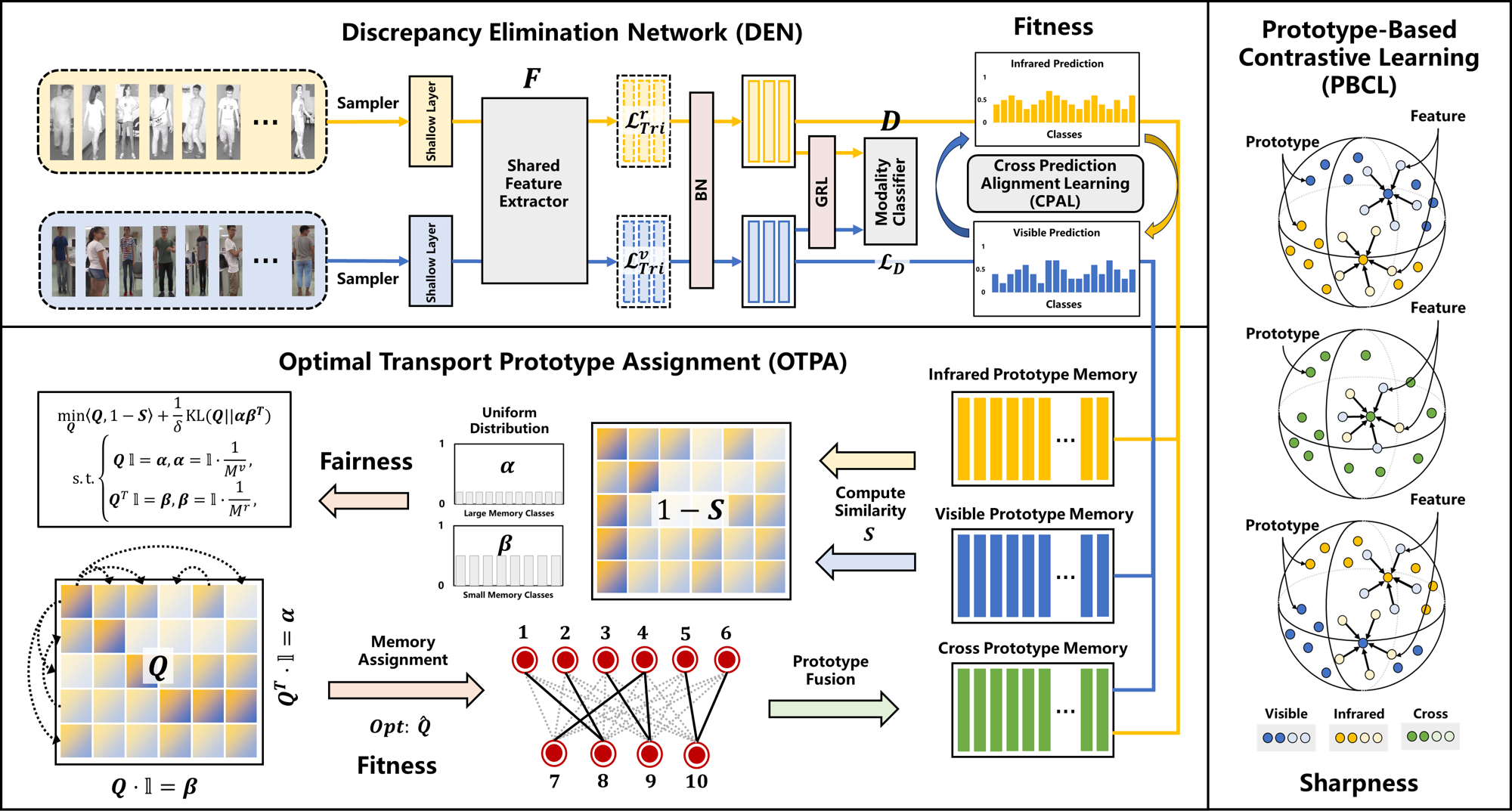}
	\caption{The pipeline of our framework. Discrepancy Elimination Network (DEN) extracts robust features for unlabeled data. After that, we use DBSCAN to generate clustering pseudo labels of each modality and create visible and infrared prototype memories. The Optimal Transport Prototype Assignment (OTPA) is then proposed to find the cross-modality correspondence. This correspondence helps us to obtain the cross-modality pseudo labels and create the cross prototype memory. Based on these memories, we design various Prototype-Based Contrastive Learning losses (PBCL), including VCL, ICL, CCL, and MCL, to minimize intra- and cross-modality entropy. We also use a Cross Prediction Alignment Learning (CPAL) referring to OTLA-ReID \cite{OTLA-ReID} to reduce negative effects brought by inaccurately matched cross-modality data.}
	\label{Fig:main}
\end{figure*}

\textbf{Learning Framework.} Fig. \ref{Fig:main} shows our designed learning framework. Our feature extraction network is followed by OTLA-ReID \cite{OTLA-ReID}, {\it i.e.,} discrepancy elimination network (DEN). Then optimal transport prototype assignment algorithm (OTPA) is designed to find the cross-modality correspondence among clustering prototypes. To overcome all prototypes being grouped into one, we add the uniform prior to our optimal transport objective function. Besides, to reach "Sharpness", we design prototype-based contrastive learning losses (PBCL) as an equivalent substitute for entropy minimization.%Finally, we also propose a cross prediction alignment learning loss (CPAL) to further filter out the negative effect brought by wrongly matched cross-modality data, which can be a part of "Fitness".} 
Next, we will elaborate on each component and illustrate how they cooperate with each other under the guidance of learning principles, {\it i.e.,} ''Sharpness" (entropy minimization), ''Fairness" (uniform distribution) and ''Fitness" (reliable cross-modality matching).

\textbf{Feature Extraction Network.}  %served as a feature extractor enabling us to learn modality-invariant features. 
We sample a batch of visible and infrared images and forward them into a ResNet-50 based backbone $\boldsymbol{F}$ for feature extraction, {\it i.e.,} $\boldsymbol{f}^{v}={\rm BN}(\boldsymbol{F}(\boldsymbol{x}^{v}))$,  $\boldsymbol{f}^{r}={\rm BN}(\boldsymbol{F}(\boldsymbol{x}^{r}))$, where ${\rm BN}(\cdot)$ denotes the batch normalization. 
After that, we deploy a modality classifier $\boldsymbol{D}$ to judge which modality the feature comes from. To align the cross-modality features, {\it i.e.,} $\boldsymbol{f}^{v}$ and $\boldsymbol{f}^{r}$,  the learning objective is thus formulated as:
\begin{equation}
\begin{aligned}
	\mathcal{L}_{D} =\max_{\boldsymbol{F}} \min_{\boldsymbol{D}} \mathbb{E}_{\boldsymbol{f}^{v}}[\log(1-\boldsymbol{D}(\boldsymbol{f}^{v})]+\mathbb{E}_{\boldsymbol{f}^{r}}[\log(\boldsymbol{D}(\boldsymbol{f}^{r})]. \label{modality_discriminator_loss}
\end{aligned}
\end{equation}
Note that Eq. (\ref{modality_discriminator_loss}) can be optimized by a gradient reversal layer (GRL) \cite{GRL} technique. It acts as an identity transform during the forward propagation, while flipping the gradient of the modality classifier with a factor of $-\gamma$ in the backpropagation.  As a result, GRL could efficiently optimize Eq.(\ref{modality_discriminator_loss}), which encourages the cross-modality features ({\it i.e.,} $\boldsymbol{f}^{v}$ and $\boldsymbol{f}^{r}$) to be indistinguishable, and therefore allows for extracting modality-invariant features.

\subsection{Optimal Transport Prototype Assignment (OTPA)}
After extracting features, "Fitness" requires us to find the matched cross-modality data pairs containing the same identity, {\it i.e.,} cross-modality correspondence. Different from OTLA-ReID \cite{OTLA-ReID}, the correspondence in this paper is based on finding the matched visible and infrared prototypes, ensuring that the correspondence embeds robust cross-modality correlated information while maintaining a low computational cost. To this end, we first apply DBSCAN \cite{DBSCAN} (remove the clustering outlier data) to produce clustering pseudo labels for each modality independently. With the help of these cluster-aware pseudo labels, we can initialize the modality-specific prototype memories by averaging the extracted feature representations belonging to the same cluster:
\begin{equation}
\begin{aligned}
    \boldsymbol{\mathcal{M}}^{v}_{i} = \frac{1}{|\boldsymbol{D}^{v}_{i}|}\sum_{\boldsymbol{{f}}^{v} \in \boldsymbol{D}^{v}_{i}} \boldsymbol{{f}}^{v},
\end{aligned}
\end{equation}
\begin{equation}
\begin{aligned}
    \boldsymbol{\mathcal{M}}^{r}_{i} = \frac{1}{|\boldsymbol{D}^{r}_{i}|}\sum_{\boldsymbol{{f}}^{r} \in \boldsymbol{D}^{r}_{i}} \boldsymbol{{f}}^{r},
\end{aligned}
\end{equation}
where $\boldsymbol{\mathcal{M}}^{v/r}_{i}$ denotes the $i$-th prototype in visible/infrared memory. Note that the numbers of slots in two memories are different. $\boldsymbol{D}^{v/r}_{i}$ denotes $i$-th visible/infrared set from a specific cluster. $|\cdot|$ denotes the number of instances per cluster.

Thus establishing the cross-modality correspondence between each slot in  $\boldsymbol{\mathcal{M}}^{r}$ and $\boldsymbol{\mathcal{M}}^{v}$ is actually a cross-modality matching problem. In our approach, modality-specific prototype memory with a larger number of prototypes is viewed as suppliers while the other one is viewed as demanders. The goal of this task is to find a transportation plan $\boldsymbol{Q}$ that transports prototypes from suppliers to demands at the lowest transportation cost. Here the transportation cost is related to the similarities among prototypes. Thus we propose the following optimal transport objective function for solving this bipartite matching problem:
\begin{equation} 
\begin{aligned}
    \min_{\boldsymbol{Q}} &\left\langle\boldsymbol{Q}, 1 -\frac{\boldsymbol{S}+1}{2}\right\rangle
    \label{objective otpa}
\end{aligned}
\end{equation}
where $\boldsymbol{S}$ denotes the cosine similarity score matrix between the above two modality-specific prototype memories ranged from $[-1,1]$, and $\left\langle \cdot \right\rangle$ denotes the Frobenius dot-product. Minimizing Eq. (\ref{objective otpa}) implies that our transport plan favors the lower transportation cost, {\it i.e.,}  higher cross-modality similarity.

\begin{figure}[!t]
	\centering
	\includegraphics[scale=0.39]{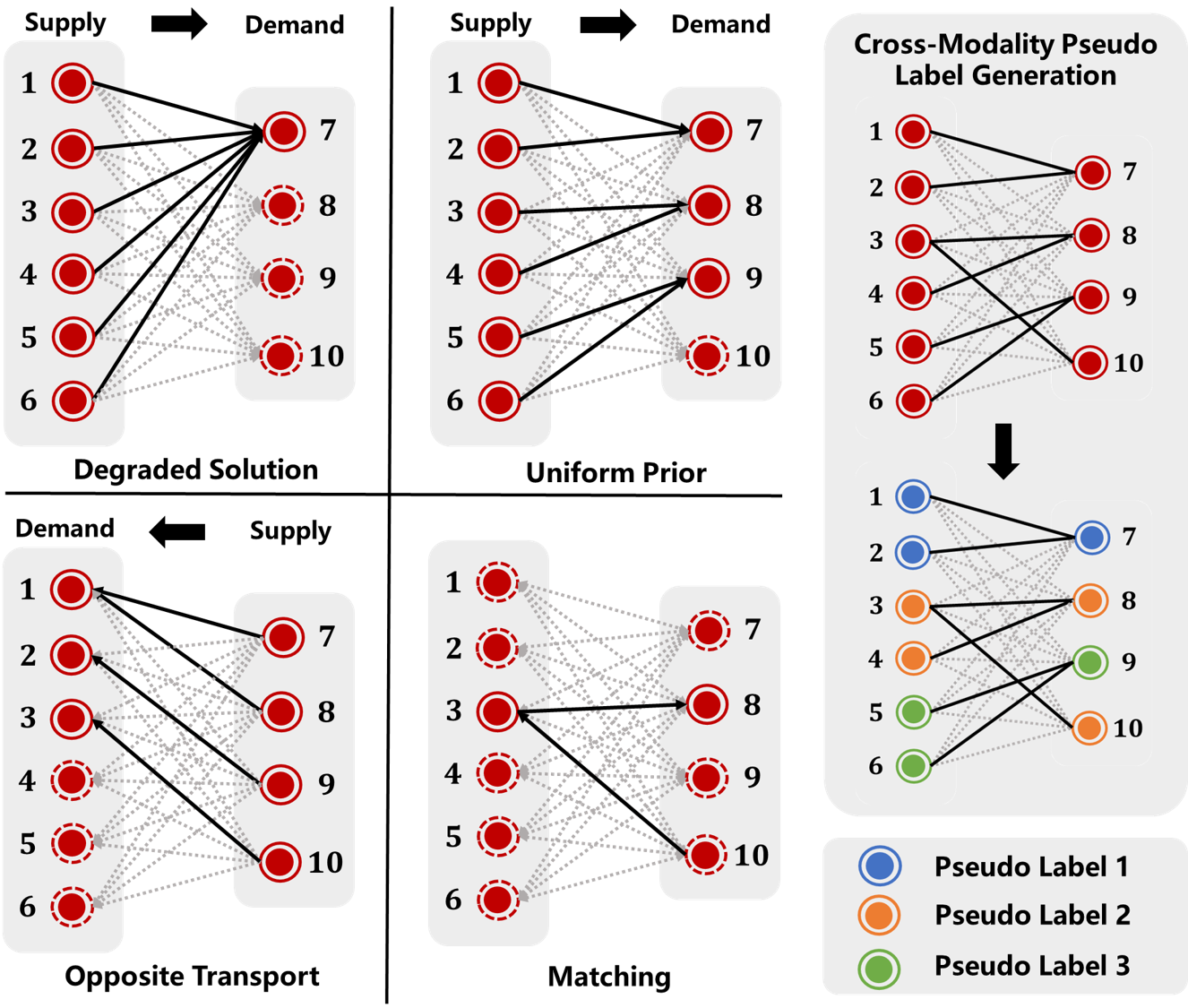}
 	\caption{Cross-Modality Pseudo Label Generation. Degraded Solution: Without uniform prior, it would meet collapsed solution. Uniform Prior: OTPA benefits from the balanced assignment with the uniform prior, but there are still unassigned demand prototypes, {\it e.g.,} node $10$. Opposite Transport: We conduct an opposite transport and find the matched nodes with respect to the unassigned demand prototypes, {\it e.g.,} node $3$. Matching: final cross-modality correspondence. }
	\label{Fig:CPLG}
\end{figure}

However, simply optimizing Eq. (\ref{objective otpa}) may lead to a collapsed solution, {\it i.e.,} most supply prototypes are transported to very few demand prototypes (Degraded Solution in Fig. \ref{Fig:CPLG}). This is because the model would not be discriminative in the initial stage, and therefore the similarity can not reflect the true matching information. To prevent this collapsed phenomenon, we enforce the "Fairness" regularization in the optimization from two intuitions: 1) each supply prototype owns an equiprobable assignment choice; 2) each demand prototype owns approximately the same number of supply prototypes.

To achieve intuitions, we can modify Eq. (\ref{objective otpa}) by incorporating two uniformly distributed constraints (suppose visible prototypes are suppliers and infrared prototypes are demanders):
\begin{equation} 
\begin{aligned}
 \min_{\boldsymbol{Q}} &\left\langle\boldsymbol{Q}, 1 -\frac{\boldsymbol{S}+1}{2}\right\rangle + \frac{1}{\delta}{\rm KL}(\boldsymbol{Q}||\boldsymbol{\alpha}\boldsymbol{\beta}^{T}).\\
 	&\text{s.t.} \left\{
	\begin{aligned}
		& \boldsymbol{Q} \mathbbm{1} = \boldsymbol{\alpha},\quad \boldsymbol{\alpha} = \mathbbm{1} \cdot \frac{1}{M^{v}},\\ & \boldsymbol{Q}^{T} \mathbbm{1} = \boldsymbol{\beta},\quad \boldsymbol{\beta} = \mathbbm{1} \cdot \frac{1}{M^{r}}, \\ & \boldsymbol{S} = {\rm Norm}(\boldsymbol{{\mathcal{M}}}^{v}){\rm Norm}(\boldsymbol{{\mathcal{M}}}^{r})^{T},
	\end{aligned}
	\right.
 \label{objective otpa final}
\end{aligned}
\end{equation}
where $\boldsymbol{\alpha} \in \mathbb{R}^{M^{v}}$ and $\boldsymbol{\beta} \in \mathbb{R}^{M^{r}}$ represent the prior uniform distributed vectors, ${M^{v}}$ and ${M^{r}}$ denote the length of visible and infrared memory respectively, ${\rm KL(\cdot)}$ denotes the KL-divergence, ${\rm Norm}(\cdot)$ denotes normalized manipulation. In essence, we use KL-divergence to constrain the transportation plan $\boldsymbol{Q}$ to uniformly transport each supply prototype to the demand prototype, leading to a "fair" matching result. $\delta$ is a trade-off hyperparameter controlling the weight of this uniform regularization. Through them, the supply prototypes are forced to be assigned to the equally-sized subsets, avoiding the collapsed solution shown in Figure \ref{Fig:CPLG} of Uniform Prior.

With the uniform constraint, we can adopt a fast version of Sinkhorn-Knopp algorithm \cite{Sinkhorn-Knopp} to solve the above optimal transport problem. In this solver, the optimal solution $\hat{\boldsymbol{Q}}$ can be achieved through the iterative matrix scaling operation:
\begin{equation}
\begin{aligned}
	\forall m: \boldsymbol{\alpha}_{m} \gets [(\boldsymbol{K}) \boldsymbol{\beta}]_{m}^{-1} \quad \forall n: \boldsymbol{\beta}_{n} \gets [\boldsymbol{\alpha}^{T}(\boldsymbol{K})]_{n}^{-1},
\end{aligned}  
\end{equation}
where $\boldsymbol{K}:=e^{-\delta(1 -\frac{\boldsymbol{S}+1}{2})}$ is the element-wise exponential of $-\delta(1 -\frac{\boldsymbol{S}+1}{2})$. Specifically, $\boldsymbol{\alpha}$ initializes with $\mathbbm{1} \cdot \frac{1}{M^{v}}$ and $\boldsymbol{\beta}$ initializes with $\mathbbm{1} \cdot \frac{1}{M^{r}}$. When the iteration meets the termination conditions or exceeds the maximum number, the auxiliary vectors $\boldsymbol{\alpha}$ and $\boldsymbol{\beta}$ are fixed. One primary advantage of this algorithm is that the optimal solution matrix of Eq. (\ref{objective otpa final}) can be equivalently converted as:
\begin{equation}
\begin{aligned}
    \hat{\boldsymbol{Q}} = \rm{diag}(\boldsymbol{\alpha}) \boldsymbol{K} \rm{diag}(\boldsymbol{\beta}),
\end{aligned}
\end{equation}
where $\rm{diag}(\cdot)$ denotes the square diagonal matrix with the elements of the corresponding vector on the main diagonal.

The maximal value of the optimal transport plan $\hat{\boldsymbol{Q}}$ indicates the prototype-level cross-modality correspondence. This correspondence is then acted as an important semantic monitor for training the model. However, the number of visible prototypes $M^{v}$ is different from infrared ones $M^{r}$, leading to a one-to-many matching. Besides, though with uniform regularization, some demand prototypes are still not assigned. To deal with this issue, we show our implementation as a toy example in Fig. \ref{Fig:CPLG}. If there are unassigned demander {\it e.g.,} node $10$, we would conduct an opposite transport by exchanging suppliers and demanders. In this case, node $10$ is transported to node $3$ and they are considered to be matched. Finally, we merge the nodes with cross-modality correspondence and treat them as a single class, {\it e.g.,} node $3$, $4$, $8$, $10$ share the same cross-modality pseudo label in this process.

After that, we obtain the cross-modality pseudo labels and then can construct a cross-modality prototype memory $\boldsymbol{\mathcal{M}}^{c}$. $\boldsymbol{\mathcal{M}}^{c}$ is initialized by averaging the cross-modality features with the same generated cross-modality pseudo labels. Note that we now have three kinds of prototype memories (\textit{i.e.}, visible prototype memory, infrared prototype memory and cross prototype memory) and we will initialize them at the beginning of each training epoch. We use the momentum updating strategy to update the prototype memories with coefficient $m$ at every training batch:
\begin{equation}
\begin{aligned}
    \boldsymbol{\mathcal{M}}^{v}_{i} = m \cdot \boldsymbol{\mathcal{M}}^{v}_{i} + (1 - m) \cdot \boldsymbol{{f}}^{v},
\end{aligned}
\label{memory update 1}
\end{equation}
\begin{equation}
\begin{aligned}
    \boldsymbol{\mathcal{M}}^{r}_{i} = m \cdot \boldsymbol{\mathcal{M}}^{r}_{i} + (1 - m) \cdot \boldsymbol{{f}}^{r},
\end{aligned}
\label{memory update 2}
\end{equation}
\begin{equation}
\begin{aligned}
    \boldsymbol{\mathcal{M}}^{c}_{i} = m \cdot \boldsymbol{\mathcal{M}}^{c}_{i} + (1 - m) \cdot \frac{\boldsymbol{\hat{f}}^{v} + \boldsymbol{\hat{f}}^{r}}{2},
\end{aligned}
\label{memory update 3}
\end{equation}
where $\boldsymbol{{f}}^{v/r}$ denotes the visible or infrared feature representation in the current training batch from the $i$-th cluster. $\boldsymbol{\hat{f}}^{v}$ and $\boldsymbol{\hat{f}}^{r}$ are shared with the same cross-modality pseudo label. They are used to update the corresponding cross prototype. As we can see, modality-specific prototype memory  $\boldsymbol{\mathcal{M}}^{v/r} $ is updated by the intra-modality clustering-aware relationship, while the cross-modality prototype memory $\boldsymbol{\mathcal{M}}^{c} $ is updated by the learned cross-modality correspondence.

\subsection{Prototype-Based Contrastive Learning (PBCL)}
\begin{figure}[!t]
	\centering
	\includegraphics[scale=0.31]{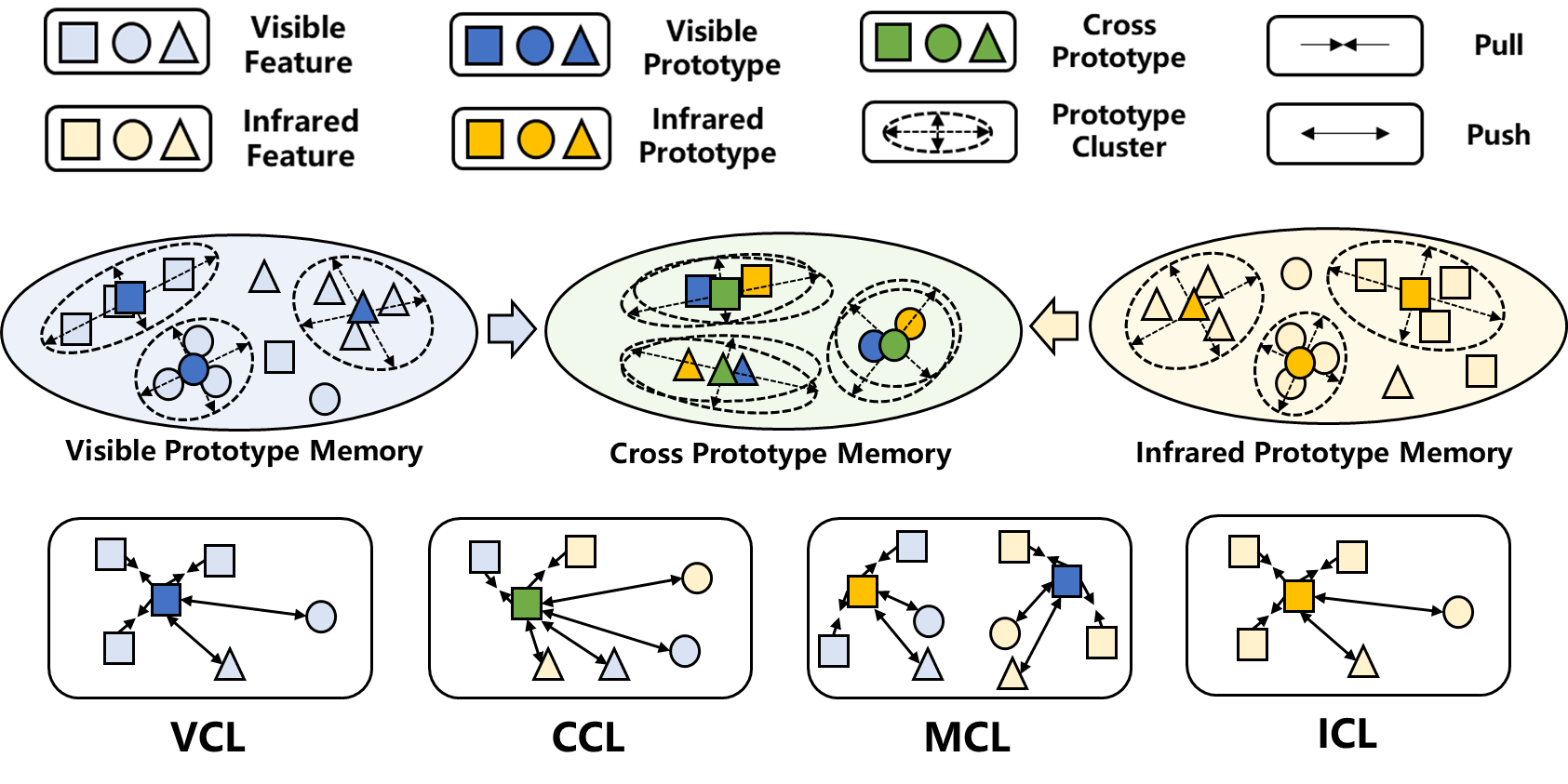}
	\caption{The diagram of prototype-based contrastive learning losses, including VCL, ICL, CCL, and MCL. These losses can be the substitutes for entropy minimization due to the lack of a proper classifier.} 
	\label{Fig:PBCL}
\end{figure}

Based on the established prototype memories, we propose various prototype-based contrastive learning losses shown in Fig. \ref{Fig:PBCL}, as an alternative to entropy minimization suggested by "Sharpness".

\textbf{Modality-Specific Contrastive Learning (VCL, ICL).} Modality-specific contrastive learning consists of two parts: visible contrastive learning (VCL) and infrared contrastive learning (ICL). These two kinds of contrastive learning losses are used to minimize the intra-modality entropy and therefore promote discrimination. VCL and ICL are defined as:
\begin{equation}
\begin{aligned}
    \mathcal{L}_{VCL} = - \frac{1}{V} \sum^{V}_{k=1} {\rm log}\frac{{\rm exp}(\boldsymbol{{f}}^{v}_{k} \cdot \boldsymbol{{\mathcal{M}}}^{v}_{+} / \tau)}{\sum_{j} {\rm exp}(\boldsymbol{f}^{v}_{k} \cdot \boldsymbol{{\mathcal{M}}}^{v}_{j} / \tau)},
\end{aligned}
\end{equation}
\begin{equation}
\begin{aligned}
    \mathcal{L}_{ICL} = - \frac{1}{R} \sum^{R}_{k=1} {\rm log}\frac{{\rm exp}(\boldsymbol{{f}}^{r}_{k} \cdot \boldsymbol{{\mathcal{M}}}^{r}_{+} / \tau)}{\sum_{j} {\rm exp}(\boldsymbol{f}^{r}_{k} \cdot \boldsymbol{{\mathcal{M}}}^{r}_{j} / \tau)},
\end{aligned}
\end{equation}
where $V/R$ denotes the number of visible/infrared samples in each training batch,  $\boldsymbol{{\mathcal{M}}}^{v/r}_{+}$ denotes the positive modality-specific prototype corresponding to the clustering-aware pseudo label of the $k$-th query feature $\boldsymbol{f}_{k}$, and $\tau$ denotes the temperature hyper-parameter.

\textbf{Cross Contrastive Learning (CCL).} With the cross-modality correspondence acquired from OTPA, we can also compute prototype-based contrastive learning loss to minimize the cross-modality entropy:
\begin{equation}
\begin{aligned}
    \mathcal{L}_{CCL} = - \frac{1}{V+R} \sum_{k=1}^{V+R}{\rm log}\frac{{\rm exp}(\boldsymbol{{f}}^{v/r}_{k} \cdot \boldsymbol{{\mathcal{M}}}^{c}_{+} / \tau)}{\sum_{j}{\rm exp}(\boldsymbol{{f}}^{v/r}_{k} \cdot \boldsymbol{{\mathcal{M}}}^{c}_{j} / \tau)},
\end{aligned}
\end{equation}
where $\boldsymbol{\hat{\mathcal{M}}}^{c}_{+}$ denotes the positive cross prototype
corresponding to the cross-modality pseudo label of $k$-th query feature $\boldsymbol{f}_{k}$. 

\textbf{Mutual Contrastive Learning (MCL).} To further minimize the cross-modality entropy, we dig up the correlations between the above two kinds of pseudo labeling frameworks, {\it i.e.,} intra-modality clustering-aware pseudo labels and cross-modality pseudo labels generated from OPTA, so as to construct prototype-based contrastive learning loss. In this loss, for each sample in the training batch, we first compute the cosine similarity between its feature representation and the modality-specific prototypes of the opposite modality. Taking the visible query feature as an example, we can find the cosine similarity as: 
\begin{equation}
\begin{aligned}
    s_{ki}^{v} = {\rm Norm}(\boldsymbol{f}^{v}_k)^{T}{\rm Norm}(\boldsymbol{\hat{\mathcal{M}}}_{i}^{r}),
\end{aligned}
\end{equation}
where $\boldsymbol{\hat{\mathcal{M}}}^{r}$ denotes a subset of ${\boldsymbol{\mathcal{M}}}^{r}$ and ${\rm Norm}(\cdot)$ denotes normalized manipulation. To construct $\boldsymbol{\hat{\mathcal{M}}}^{r}$, we select the prototypes in ${\boldsymbol{\mathcal{M}}}^{r}$ sharing the same cross-modality pseudo label when given a query visible sample. Note that the prototypes in $\boldsymbol{\hat{\mathcal{M}}}^{r}$ may come from one or several kinds of DBSCAN clusters. Then if this cosine similarity score exceeds a threshold ($\theta=0.2$), we treat them as the positive similarity scores $\hat{s}_{ki}^{v}$. Thus, the loss function can be formulated as:
\begin{equation}
\begin{aligned}
    \mathcal{L}_{MCL} & = - \frac{1}{V} \sum_{k=1}^{V} {\rm log}\frac{\sum_{i} {\rm exp}(\hat{s}_{ki}^{v} / \tau)}{\sum_{j} {\rm exp}(\boldsymbol{{f}}^{v}_{k} \cdot \boldsymbol{{\mathcal{M}}}_{j}^{r} / \tau)} \\ 
    & - \frac{1}{R} \sum_{k=1}^{R} {\rm log}\frac{\sum_{i} {\rm exp}(\hat{s}_{ki}^{r} / \tau)}{\sum_{j} {\rm exp}(\boldsymbol{{f}}^{r}_{k} \cdot \boldsymbol{{\mathcal{M}}}_{j}^{v} / \tau)}.
\end{aligned}
\end{equation}

\subsection{Optimization}
The total training process can be divided into two stages, \textit{i.e.}, \textbf{pretraining stage} and \textbf{our proposed training stage}. As for pretrainig stage, we pretrain the model like PGM \cite{PGM} using the following loss functions:
\begin{equation}
\begin{aligned}
    \mathcal{L} =  \mathcal{L}_{VCL} + \mathcal{L}_{ICL} + \lambda\mathcal{L}_{Tri}.
\end{aligned}
\end{equation}

As for our proposed training stage, we further train the model using the complete loss functions below based on the pretraining weights:
\begin{equation}
\begin{aligned}
    \mathcal{L} = & \mathcal{L}_{D} + \lambda\mathcal{L}_{Tri} + \mathcal{L}_{VCL} + \mathcal{L}_{ICL} \\ & + \mathcal{L}_{CCL} + \mathcal{L}_{MCL} + \mathcal{L}_{CPAL},
\end{aligned}
\end{equation}
where $\mathcal{L}_{Tri}$ denotes the intra-modality triplet loss by giving cross-modality pseudo labels generated by OTPA, $\mathcal{L}_{CPAL}$ denotes the cross prediction alignment learning loss (CPAL), which has a similar form as prediction alignment learning loss proposed in OTLA-ReID \cite{OTLA-ReID} (the prediction here is defined as the cosine similarities between the query features and the prototypes in the cross prototype memory, where more details will be shown in Appendix B.1). The  $\lambda = 0.5$ is a hyper-parameter. Other weights of loss functions are simply set to 1.0.

\section{Experiment}
% In this section, we conduct extensive experiments to provide a comprehensive evaluation of our proposed method.

\subsection{Experimental Settings}
\subsubsection{Datasets Summary} 
The proposed method has been evaluated on two widely adopted benchmarks \textbf{SYSU-MM01} \cite{SYSU-MM01} and \textbf{RegDB} \cite{RegDB}. SYSU-MM01 is a large-scale cross-modality ReID dataset that is collected by four RGB and two near-infrared cameras from both indoor and outdoor environments. It's composed of 287,628 visible images and 15,792 infrared images in total of  491 identities. RegDB is collected by two aligned cameras (one visible and one far-infrared). It includes 412 identities and each identity has 10 infrared images and 10 visible images. Hence RegDB totally contains 4,120 visible images and 4,120 far-infrared images.

\subsubsection{Evaluation Metrics}
On both datasets, we follow the popular protocols \cite{DDAG} for evaluation, in which cumulative match characteristic (CMC), mean average precision (mAP), and mean inverse negative penalty (mINP) \cite{AGW} are adopted. SYSU-MM01 contains two different testing settings, {\textit{i.e., All Search}} mode and {\textit{Indoor Search}} mode. For all-search mode, the gallery set consists of all visible images (captured by CAM1, CAM2, CAM4, CAM5), and the query set is composed of all infrared samples (captured by CAM3, CAM6). For indoor-search mode, images are captured only from the indoor scene to constitute the gallery set, excluding the images from CAM4 and CAM5. On both search modes, the proposed method is evaluated under the single-shot setting. For RegDB, we randomly select 206 identities for training and the remaining 206 identities for testing. It also includes two different settings, \textit{i.e.}, \textit{Visible to Thermal} mode and \textit{Thermal to Visible} mode. We report the average result by randomly splitting of training and testing set over 10 times.

\renewcommand\arraystretch{1.1}
\begin{table*}[h]
	\centering
	\caption{Comparisons with SOTA methods on SYSU-MM01 of \textit{All Search} mode and \textit{Indoor Search} mode, including unsupervised ReID  (USL-ReID), fully-supervised VI-ReID (SVI-ReID) and unsupervised VI-ReID (USVI-ReID). All methods are measured by CMC(\%), mAP(\%) and mINP(\%). $^\dagger$ indicates we re-implement the result with the official code.}
	\resizebox{2.0\columnwidth}{!}{
		\begin{tabular}{c|c|c|c|c|c|c|c|c|c|c|c|c}
			\toprule[1pt]
            \hline
			\multicolumn{3}{c|}{Settings} &
			\multicolumn{5}{c|}{All Search} &
			\multicolumn{5}{c}{Indoor Search} \\
			\hline
			Type & Method & Venue & Rank-1 & Rank-10 & Rank-20 & mAP & mINP & Rank-1 & Rank-10 & Rank-20 & mAP & mINP \\
			\hline
            % \midrule[1pt]
			% \multirow{7}{*}{UDA-ReID} & SSG\cite{SSG} & ICCV'\textcolor{blue}{19} & 2.3 & 17.2 & 28.9 & 5.0 & - & - & - & - & - & - \\
			% & ECN\cite{ECN} & CVPR'\textcolor{blue}{19} & 8.1 & 32.5 & 45.9 & 12.7 & - & - & - & - & - & -  \\
			% & D-MMD$^\dagger$\cite{D-MMD} & ECCV'\textcolor{blue}{20} & 12.5 & 29.2 & 37.6 & 10.4 & - & 19.0 & 39.5 & 46.5 & 15.4 & - \\
			% & MMT$^\dagger$\cite{MMT} & ICLR'\textcolor{blue}{20} & 13.9 & 34.0 & 41.4 & 8.4 & - & 21.0 & 44.6 & 53.9 & 15.3 & - \\
			% & SpCL(UDA)$^\dagger$\cite{SpCL} & NIPS'\textcolor{blue}{20} & 15.1 & 42.7 & 52.3 & 6.5 & - & 19.5 & 51.4 & 60.7 & 12.1 & - \\
			% & GLT$^\dagger$\cite{GLT} & CVPR'\textcolor{blue}{21} & 7.7 & 33.4 & 45.3 & 9.5 & - & 12.1 & 44.6 & 61.4 & 18.0 & - \\
   %          & SECRET$^\dagger$\cite{SECRET} & AAAI'\textcolor{blue}{22} & 6.6 & 24.3 & 34.8 & 9.4 & - & 9.3 & 38.0 & 54.5 & 15.8 & - \\
			% \hline
			\multirow{7}{*}{USL-ReID} & BUC$^\dagger$\cite{BUC} & AAAI'\textcolor{blue}{19} & 8.2 & 22.7 & 31.1 & 3.2 & - & 12.5 & 27.9 & 36.6 & 6.0 & - \\
			& SpCL(USL)$^\dagger$\cite{SpCL} & NIPS'\textcolor{blue}{20} & 18.7 & 39.4 & 47.5 & 11.4 & - & 27.1 & 49.9 & 58.7 & 20.9 & - \\
			& MetaCam$^\dagger$\cite{MetaCam} & CVPR'\textcolor{blue}{21} & 14.7 & 32.7 & 40.5 & 9.3 & - & 23.9 & 43.8 & 51.6 & 17.1 & - \\
			& HCD$^\dagger$\cite{HCD} & ICCV'\textcolor{blue}{21} & 18.0 & 48.8 & 60.9 & 17.9 & - & 24.4 & 61.9 & 76.1 & 28.8 & - \\
            & ICE\cite{ICE} & ICCV'\textcolor{blue}{21} & 20.5 & 57.5 & 70.9 & 10.2 & 20.4 & 29.8 & 69.4 & 82.7 & 38.4 & 34.3 \\
            & STS$^\dagger$\cite{STS} & TIP'\textcolor{blue}{22} & 18.7 & 51.8 & 63.9 & 19.9 & - & 26.8 & 77.5 & 31.3 & - \\
            & PPLR$^\dagger$\cite{PPLR} & CVPR'\textcolor{blue}{22} & 19.1 & 54.1 & 66.4 & 18.6 & - & 20.6 & 58.5 & 71.9 & 26.3 & - \\
            & DCMIP$^\dagger$\cite{DCMIP} & ICCV'\textcolor{blue}{23} & 21.3 & 58.7 & 71.9 & 19.5 & - & 25.5 & 65.9 & 84.2 & 40.2 & - \\
			\hline
			\multirow{9}{*}{SVI-ReID} 
			& JSIA-ReID\cite{JSIA-ReID} & AAAI'\textcolor{blue}{20} & 38.1 & 80.7 & 89.9 & 36.9 & - & 43.8 & 86.2 & 94.2 & 52.9 & - \\
			& Hi-CMD\cite{Hi-CMD} & CVPR'\textcolor{blue}{20} & 34.9 & 77.6 & - & 35.9 & - & - & - & - & - & - \\
			& AGW\cite{AGW} & TPAMI'\textcolor{blue}{21} & 47.5 & 84.4 & 92.1 & 47.7 & 35.3 & 54.2 & 91.1 & 96.0 & 63.0 & 59.2 \\
			% & NFS\cite{NFS} & CVPR'\textcolor{blue}{21} & 56.9 & 91.3 & 96.5 & 55.5 & - & 62.8 & 96.5 & 99.1 & 69.8 & - \\
            & MPANet\cite{MPANet} & CVPR'\textcolor{blue}{21} & 70.6 & 96.2 & 98.8 & 68.2 & - & 76.7 & 98.2 & 99.6 & 81.0 & - \\
			& LbA\cite{LbA} & ICCV'\textcolor{blue}{21} & 55.4 & - & - & 54.1 & - & 58.5 & - & - & 66.3 & - \\
			% & CAJL\cite{CAJL} & ICCV'\textcolor{blue}{21} & 69.9 & 95.7 & 98.5 & 66.9 & - & 76.3 & 97.9 & 99.5 & 80.4 & - \\
            & FMCNet\cite{FMCNet} & CVPR'\textcolor{blue}{22} & 66.3 & - & - & 62.5 & - & 68.2 & - & - & 74.1 & - \\
            % & CMT\cite{CMT} & ECCV'\textcolor{blue}{22} & 71.9 & 96.5 & 98.9 & 68.6 & - & 76.9 & 97.7 & 99.6 & 79.9 & - \\
            & MSCLNet\cite{MSCLNet} & ECCV'\textcolor{blue}{22} & 77.0 & 97.6 & 99.2 & 71.6 & - & 78.5 & 99.3 & 99.9 & 81.2 & - \\
            & SEFL\cite{SEFL} & CVPR'\textcolor{blue}{23} & 77.1 & 97.0 & 99.1 & 72.3 & - & 82.1 & 97.4 & 98.9 & 83.0 & - \\
            & PartMix\cite{PartMix} & CVPR'\textcolor{blue}{23} & 77.8 & - & - & 74.6 & - & 81.5 & - & - & 84.4 & - \\
			\hline
			\multirow{9}{*}{USVI-ReID}
			& H2H\cite{H2H} & TIP'\textcolor{blue}{21} & 25.5 & 63.9 & 76.2 & 25.2 & - & 13.9 & 30.4 & 39.3 & 12.7 & - \\
            & OTLA-ReID\cite{OTLA-ReID} & ECCV'\textcolor{blue}{22} & 29.9 & 71.8 & 83.9 & 27.1 & - & 29.8 & 74.1 & 87.6 & 38.8 & - \\
            & ADCA\cite{ADCA} & ACM MM'\textcolor{blue}{22} & 45.5 & 85.3 & 93.2 & 42.7 & 28.3 & 50.6 & 89.7 & 96.2 & 59.1 & 55.2 \\
            & CHCR\cite{CHCR} & TCSVT'\textcolor{blue}{23} & 47.7 & 87.3 & 94.2 & 45.3 & - & 50.1 & 90.8 & 95.6 & 42.2 & - \\
            & DOTLA\cite{DOTLA} & ACM MM'\textcolor{blue}{23} & 50.4 & 89.0 & 95.9 & 47.4 & 32.4 & 53.5 & 92.2 & 97.8 & 61.7 & 57.4 \\
            & MBCCM\cite{MBCCM} & ACM MM'\textcolor{blue}{23} & 53.1 & 89.6 & 96.7 & 48.2 & 32.4 & 55.2 & 91.4 & 95.8 & 62.0 & 57.1 \\
            & CCLNet\cite{CCLNet} & ACM MM'\textcolor{blue}{23} & 54.0 & 88.8 & 95.0 & 50.2 & - & 56.7 & 91.1 & 97.2 & 65.1 & - \\
            & PGM\cite{PGM} & CVPR'\textcolor{blue}{23} & 57.3 & 92.5 & 97.2 & 51.8 & 35.0 & 56.2 & 90.2 & 95.4 & 62.7 & 58.1 \\
			& \textbf{OTPA-ReID (Ours)} & - & \textcolor{red}{\textbf{60.6}} & \textcolor{red}{\textbf{92.6}} & \textcolor{red}{\textbf{97.4}} & \textcolor{red}{\textbf{56.1}} & \textcolor{red}{\textbf{40.3}} & \textcolor{red}{\textbf{63.2}} & \textcolor{red}{\textbf{92.6}} & \textcolor{red}{\textbf{97.0}} & \textcolor{red}{\textbf{68.3}} & \textcolor{red}{\textbf{63.5}} \\
			\hline
            \bottomrule[1pt]
		\end{tabular}
	}
	\label{Tab:SYSU-MM01}
\end{table*}

\subsection{Implementation Details}
\subsubsection{Training Details}
We implement our method using PyTorch and MindSpore on one NVIDIA RTX3090Ti GPU. The batch size is fixed to 64 for all experiments. According to the cross-modality pseudo labels from OTPA, in every batch, we sample 4 different identities (pseudo labels), and each identity includes 8 visible images and 8 infrared images. The model is optimized by Adam optimizer \cite{Adam} with an initial learning rate of 3.5$\times$10$^{-3}$. The learning rate is incorporated with a warm-up strategy \cite{WarmUp} and decays 10 times at the $20$-th and the $50$-th epoch. The total training epochs are set to 90 (the pretraining epochs are set to 40). All the pedestrian images are resized to 288$\times$144. The margin $\rho$ of triplet loss is set to 0.3, and the $\delta$ of OTPA is fixed at 25. The temperature $\tau$ is equal to 0.05. Besides, the input images are randomly flipped and erased with 50\% probability, while visible images are extra randomly augmented by grayscale with 50\% probability. The momentum coefficient $m$ in memory update is set to 0.1 for both SYSU-MM01 and RegDB.

\subsubsection{Critical Architectures}
We adopt ResNet-50 \cite{ResNet} pretrained on ImageNet \cite{Imagenet} as our backbone. The first shallow block is modality-independent and the remaining blocks are shared. The modality classifier in DEN is implemented with three FC layers and one BN layer \cite{BN}. GRL \cite{GRL} is a non-parametric module and $\gamma=2/(1+\exp(-\eta\frac{iter}{maxiter}))-1$ controls the scale of the reversed gradient, where $\eta$ is fixed to 10, $maxiter$ is set to 10000, and $iter$ linearly increases as the training goes on.

\subsection{Main Results}
To demonstrate the effectiveness of our method, we compare our approach with four related ReID settings, including fully-supervised VI-ReID (SVI-ReID), unsupervised VI-ReID (USVI-ReID), and unsupervised learning ReID (USL-ReID). As for USL-ReID, it's designed for the visible modality unsupervised task. Here, we use both unlabeled visible and infrared data to re-implement methods of this setting, and therefore those methods would naturally perform poorly compared with other settings. The main results are shown in Tab. \ref{Tab:SYSU-MM01} and \ref{Tab:RegDB}. 

% To demonstrate the effectiveness of our method, we compare our approach with four related ReID settings, including fully-supervised VI-ReID (SVI-ReID), unsupervised VI-ReID (USVI-ReID), unsupervised domain adaptation ReID (UDA-ReID) and unsupervised learning ReID (USL-ReID). In particular, UDA-ReID and USL-ReID are designed for single-modality (RGB) ReID tasks. To make a fair comparison, for UDA-ReID, we use ground-truth labeled visible data as the source domain and unlabeled infrared data as the target domain followed by OTLA-ReID \cite{OTLA-ReID}, where infrared data without color information is more difficult to annotate. For USL-ReID, we use both unlabeled visible and infrared data to train the model and therefore those methods would naturally perform poorly compared with other settings. The main results are shown in Tab. \ref{Tab:SYSU-MM01} and \ref{Tab:RegDB}. 

\renewcommand\arraystretch{1.1}
\begin{table*}[h]
	\centering
	\caption{Comparisons with SOTA methods on RegDB of \textit{Visible to Thermal} mode and \textit{Thermal to Visible} mode, including unsupervised ReID (USL-ReID), fully-supervised VI-ReID (SVI-ReID) and unsupervised VI-ReID (USVI-ReID). All methods are measured by CMC(\%), mAP(\%) and mINP(\%). $^\dagger$ indicates we re-implement the result with the official code.}
	\resizebox{2.0\columnwidth}{!}{
		\begin{tabular}{c|c|c|c|c|c|c|c|c|c|c|c|c}
            \toprule[1pt]
			\hline
			\multicolumn{3}{c|}{Settings} &
			\multicolumn{5}{c|}{Visible to Thermal} &
			\multicolumn{5}{c}{Thermal to Visible} \\
			\hline
			Type & Method & Venue & Rank-1 & Rank-10 & Rank-20 & mAP & mINP & Rank-1 & Rank-10 & Rank-20 & mAP & mINP \\
			\hline
			% \multirow{7}{*}{UDA-ReID} & SSG\cite{SSG} & ICCV'\textcolor{blue}{19} & 1.9 & 5.1 & 7.5 & 3.2 & - & - & - & - & - & - \\
			% & ECN\cite{ECN} & CVPR'\textcolor{blue}{19} & 2.2 & 8.4 & 12.6 & 2.9 & - & - & - & - & - & - \\
			% & D-MMD$^\dagger$\cite{D-MMD} & ECCV'\textcolor{blue}{20} & 2.2 & 6.7 & 12.1 & 3.7 & - & 2.0 & 4.2 & 7.3 & 3.6 & - \\
			% & MMT$^\dagger$\cite{MMT} & ICLR'\textcolor{blue}{20} & 5.3 & 9.7 & 14.0 & 7.1 & - & 11.0 & 20.6 & 27.1 & 12.1 & - \\
			% & SpCL(UDA)$^\dagger$\cite{SpCL} & NIPS'\textcolor{blue}{20} & 3.3 & 6.5 & 9.6 & 4.3 & - & 8.4 & 19.1 & 25.4 & 9.5 & - \\
			% & GLT$^\dagger$\cite{GLT} & CVPR'\textcolor{blue}{21} & 2.9 & 4.5 & 8.3 & 4.5 & - & 6.3 & 13.9 & 21.7 & 7.6 & - \\
   %          & SECRET$^\dagger$\cite{SECRET} & AAAI'\textcolor{blue}{22} & 6.4 & 11.7 & 15.0 & 8.3 & - & 2.5 & 6.4 & 10.2 & 5.4 & - \\
			% \hline
			\multirow{7}{*}{USL-ReID} & BUC$^\dagger$\cite{BUC} & AAAI'\textcolor{blue}{19} & 4.7 & 16.9 & 24.1 & 4.5 & - & 8.8 & 25.1 & 32.9 & 6.0 & - \\
			& SpCL(USL)$^\dagger$\cite{SpCL} & NIPS'\textcolor{blue}{20} & 20.6 & 40.6 & 50.0 & 17.3 & - & 19.0 & 40.5 & 49.4 & 16.6 & - \\
			& MetaCam$^\dagger$\cite{MetaCam} & CVPR'\textcolor{blue}{21} & 23.1 & 44.6 & 55.3 & 17.5 & - & 20.9 & 43.3 & 50.8 & 16.5 & - \\
			& HCD$^\dagger$\cite{HCD} & ICCV'\textcolor{blue}{21} & 10.8 & 20.6 & 27.2 & 12.3 & - & 12.4 & 26.5 & 33.9 & 13.7 & - \\
            & ICE\cite{ICE} & ICCV'\textcolor{blue}{21} & 13.0 & 25.9 & 34.4 & 15.6 & 11.9 & 12.2 & 25.7 & 34.9 & 14.8 & 10.6 \\
            & STS $^\dagger$\cite{STS} & TIP'\textcolor{blue}{22} & 11.7 & 25.8 & 32.8 & 14.9 & - & 12.5 & 25.1 & 32.5 & 13.3 & - \\
            & PPLR$^\dagger$\cite{PPLR} & CVPR'\textcolor{blue}{22} & 12.4 & 26.2 & 34.2 & 13.2 & - & 8.6 & 20.7 & 28.8 & 10.7 & - \\
            & DCMIP$^\dagger$\cite{DCMIP} & ICCV'\textcolor{blue}{23} & 25.1 & 45.8 & 57.2 & 19.0 & - & 24.9 & 44.9 & 56.2 & 18.2 & - \\
			\hline
			\multirow{9}{*}{SVI-ReID} 
			& JSIA-ReID\cite{JSIA-ReID} & AAAI'\textcolor{blue}{20} & 48.5 & - & - & 49.3 & - & 48.1 & - & - & 48.9 & - \\
			& Hi-CMD\cite{Hi-CMD} & CVPR'\textcolor{blue}{20} & 70.9 & 86.4 & - & 66.0 & - & - & - & - & - & - \\
			& AGW\cite{AGW} & TPAMI'\textcolor{blue}{21} & 70.1 & 86.2 & 91.6 & 66.4 & 50.2 & 70.5 & 87.2 & 91.8 & 65.9 & 51.2 \\
			% & NFS\cite{NFS} & CVPR'\textcolor{blue}{21} & 80.5 & 91.9 & 95.1 & 72.1 & - & 77.9 & 90.5 & 93.6 & 69.8 & - \\
            & MPANet\cite{MPANet} & CVPR'\textcolor{blue}{21} & 83.7 & - & - & 80.9 & - & 82.8 & - & - & 80.7 & - \\
			& LbA\cite{LbA} & ICCV'\textcolor{blue}{21} & 74.2 & - & - & 67.6 & - & 72.4 & - & - & 65.5 & - \\
			% & CAJL\cite{CAJL} & ICCV'\textcolor{blue}{21} & 85.0 & 95.5 & 97.5 & 79.1 & - & 84.8 & 95.3 & 97.5 & 77.8 & - \\
            & FMCNet\cite{FMCNet} & CVPR'\textcolor{blue}{22} & 89.1 & - & - & 84.4 & - & 88.4 & - & - & 83.9 & - \\
            % & CMT\cite{CMT} & ECCV'\textcolor{blue}{22} & 95.2 & 98.8 & 99.5 & 87.3 & - & 91.9 & 97.9 & 99.1 & 84.5 & - \\
            & MSCLNet\cite{MSCLNet} & ECCV'\textcolor{blue}{22} & 84.2 & - & - & 81.0 & - & 83.9 & - & - & 78.3 & - \\
            & SEFL\cite{SEFL} & CVPR'\textcolor{blue}{23} & 70.2 & 86.1 & - & 52.5 & - & 67.7 & 84.7 & - & 52.3 & - \\
            & PartMix\cite{PartMix} & CVPR'\textcolor{blue}{23} & 85.7 & - & - & 82.3 & - & 84.9 & - & - & 82.5 & - \\
			\hline
			\multirow{9}{*}{USVI-ReID}
			& H2H\cite{H2H} & TIP'\textcolor{blue}{21} & 23.8 & 45.3 & 54.0 & 18.9 & - & - & - & - & - & - \\
            & OTLA-ReID\cite{OTLA-ReID} & ECCV'\textcolor{blue}{22} & 32.9 & 54.2 & 62.5 & 29.7 & - & 32.1 & 56.1 & 64.9 & 28.6 & - \\
            & ADCA\cite{ADCA} & ACM MM'\textcolor{blue}{22} & 67.2 & 82.0 & 87.4 & 64.1 & 52.7 & 68.5 & 83.2 & 88.0 & 63.8 & 49.6 \\
            & CHCR\cite{CHCR} & TCSVT'\textcolor{blue}{23} & 68.2 & 81.5 & 88.7 & 63.8 & - & 69.1 & 83.7 & 88.2 & 64.0 & - \\
            & DOTLA\cite{DOTLA} & ACM MM'\textcolor{blue}{23} & 85.6 & 94.1 & 95.5 & 76.7 & 61.6 & 82.9 & 92.3 & 94.9 & 75.0 & 58.6 \\
            & MBCCM\cite{MBCCM} & ACM MM'\textcolor{blue}{23} & 83.8 & 95.8 & 97.8 & 77.9 & 65.0 & 82.8 & 95.7 & 96.9 & 76.7 & 61.7 \\
            & CCLNet\cite{CCLNet} & ACM MM'\textcolor{blue}{23} & 69.9 & - & - & 65.5 & - & 70.2 & - & - & 66.7 & - \\
            & PGM\cite{PGM} & CVPR'\textcolor{blue}{23} & 69.5 & - & - & 65.4 & - & 69.9 & - & - & 65.2 & - \\
			& \textbf{OTPA-ReID (Ours)} & - & \textcolor{red}{\textbf{90.3}} & \textcolor{red}{\textbf{96.8}} & \textcolor{red}{\textbf{98.3}} & \textcolor{red}{\textbf{80.7}} & \textcolor{red}{\textbf{64.3}} & \textcolor{red}{\textbf{89.0}} & \textcolor{red}{\textbf{96.0}} & \textcolor{red}{\textbf{97.6}} & \textcolor{red}{\textbf{79.5}} & \textcolor{red}{\textbf{61.3}} \\
			\hline
            \bottomrule[1pt]
		\end{tabular}
	}
	\label{Tab:RegDB}
\end{table*}

\subsubsection{Comparison with Unsupervised Methods} 
As shown in the last block of Tab. \ref{Tab:SYSU-MM01} and \ref{Tab:RegDB}, here are the key observations: 1) PGM \cite{PGM} is the currently best performing unsupervised VI-ReID method, which tries to mine the cross-modality correspondence by bipartite graph matching among the cross-modality prototypes. However, the correspondence between cross-modality prototypes is not well established in this paper. This is because bipartite graph matching can only find one-to-one correspondence (\textit{i.e.}, one prototype only can match one prototype) and cannot dig up one-to-many correspondence (\textit{i.e.}, one prototype can match several suitable prototypes). Therefore, its cross-modality learning framework cannot dig up truly matched information. In contrast, our method has overcome this limitation by transferring the cross-modality matching problem into the optimal transport problem, which brings around 4\% and 20\% mAP (Rank-1) gains on SYSU-MM01 and RegDB with simple optimization design and low computational cost (PGM requires a very large training batch size); 2) Furthermore, some unsupervised VI-ReID methods like H2H \cite{H2H} and OTLA-ReID \cite{OTLA-ReID} need an extra labeled visible dataset for auxiliary training. Compared with them, our method achieves significant gains on both datasets. 3) For other unsupervised methods designed for single-modality ReID tasks, our experimental results surpass them by a significant margin. It is somewhat unfair to directly compare them since most of them don't consider the cross-modality discrepancy. We report here to highlight the difference between the unsupervised VI-ReID task and the unsupervised ReID task.  

% \subsubsection{Comparison with UDA Methods} 
% It appears that recent state-of-the-art single-modality UDA-ReID methods cannot effectively deal with the huge modality discrepancy even though given the visible label information. We conjecture this is because most UDA-ReID methods \cite{SSG,D-MMD,MMT,GLT,SECRET} rely heavily on the labeled source domain (\textit{i.e.}, need pertaining on the source domain), which drives the model to overfit on the visible data or color information. However, our proposed method achieves great results without any real label information, which indicates that the robust cross-modality representation can also be learned by the guidance of our designed learning principles.

\subsubsection{Comparison with Fully-supervised Methods}
We provide some currently well-known fully-supervised VI-ReID results for reference. Surprisingly, our method outperforms a part of fully-supervised VI-ReID methods on both SYSU-MM01 and RegDB datasets (especially on RegDB), including JSIA-ReID \cite{JSIA-ReID}, Hi-CMD \cite{Hi-CMD}, AGW \cite{AGW} and Lba \cite{LbA}. Such phenomenon indicates that label information of both modalities could be gradually aligned by our proposed optimal-transport assignment algorithm. Thus, the unsupervised training process can be gradually closed to the supervised one. Besides, we should admit that there is still some gap between our method and SOTA fully-supervised results.

\subsection{Ablation Study}
In this subsection, we conduct the ablation study to show the effectiveness of each component in our approach. Concrete experimental results are shown in Tab. \ref{Tab:ablation_SYSU-MM01} and Tab. \ref{Tab:ablation_RegDB}. We have pretrained our model before our proposed training stage, which has drawn inspiration from the strong baseline of current USVI-ReID methods, \textit{e.g.}, ADCA \cite{ADCA} and PGM \cite{PGM}. Without OTPA algorithm, we can only use the cluster-aware pseudo labels to sample training data and compute $\mathcal{L}_{Tri}$. After adopting optimal transport prototype assignment strategy (utilize $\rm OTPA$ in tables), we use generated cross-modality pseudo labels to sample training data and compute $\mathcal{L}_{Tri}$ (incorporate $\star$ in tables) Besides, we can present a complete loss framework by adding $\mathcal{L}_{CCL}$, $\mathcal{L}_{MCL}$ and $\mathcal{L}_{CPAL}$. Order $1$ is the pretraining result. Order $2$ to Order $6$ are the fine-tuning results based on the pretraining model.

\renewcommand\arraystretch{1.1}
\begin{table*}[h]
	\centering
	\caption{Ablation study in terms of CMC(\%), mAP(\%), and mINP(\%) on SYSU-MM01. $^{\star}$ indicate that we use cross-modality pseudo labels to compute $\mathcal{L}_{Tri}$. Order 1 is the pertaining results. Order $2$ to Order $6$ are the fine-tuning results based on the pretraining model.}
	\resizebox{2.0\columnwidth}{!}{
    \begin{tabular}{c|c c c c c c c c|c|c|c|c|c|c}
    \toprule[1pt]
    \hline
    & \multicolumn{8}{c|}{Components} & \multicolumn{3}{c|}{All Search} & \multicolumn{3}{c}{Indoor Search} \\
    \cline{2-15}
    \multirow{-2}{*}{Order} & $\mathcal{L}_{VCL}$ & $\mathcal{L}_{ICL}$ & $\mathcal{L}_{Tri}$ & $\mathcal{L}_{D}$ & $\mathcal{L}_{CCL}$ & $\mathcal{L}_{MCL}$ & $\mathcal{L}_{CPAL}$ & ${\rm OTPA}$ & Rank-1 & mAP & mINP & Rank-1 & mAP & mINP \\
    \hline
    1 & $\checkmark$ & $\checkmark$ & $\checkmark$ & - & - & - & - & - & 36.30 & 35.41 & 22.14 & 40.93 & 48.92 & 44.68 \\
    2 & $\checkmark$ & $\checkmark$ & $\checkmark^{\star}$ & - & - & - & - & $\checkmark$ & 32.50 & 31.44 & 18.74 & 37.99 & 46.07 & 41.85 \\
    3 & $\checkmark$ & $\checkmark$ & $\checkmark^{\star}$ & $\checkmark$ & - & - & - & $\checkmark$ & 33.72 & 32.91 & 20.14 & 39.12 & 47.19 & 43.06 \\
    4 & $\checkmark$ & $\checkmark$ & $\checkmark^{\star}$ & $\checkmark$ & $\checkmark$ & - & - & $\checkmark$ & 48.28 & 44.93 & 29.66 & 52.99 & 59.38 & 54.87 \\
    5 & $\checkmark$ & $\checkmark$ & $\checkmark^{\star}$ & $\checkmark$ & $\checkmark$ & $\checkmark$ & - & $\checkmark$ & 58.79 & 54.34 & 38.74 & 62.22 & 67.73 & 63.32 \\
    6 & $\checkmark$ & $\checkmark$ & $\checkmark^{\star}$ & $\checkmark$ & $\checkmark$ & $\checkmark$ & $\checkmark$ & $\checkmark$ & 60.58 & 56.07 & 40.25 & 63.22 & 68.27 & 63.54 \\
    \hline
    \bottomrule[1pt]
	\end{tabular}
    }
	\label{Tab:ablation_SYSU-MM01}
\end{table*}

\renewcommand\arraystretch{1.1}
\begin{table*}[h]
	\centering
	\caption{Ablation study in terms of CMC(\%), mAP(\%), and mINP(\%) on RegDB. $^{\star}$ indicate that we use cross-modality pseudo labels to compute $\mathcal{L}_{Tri}$. Order 1 is the pertaining results. Order $2$ to Order $6$ are the fine-tuning results based on the pretraining model.}
	\resizebox{2.0\columnwidth}{!}{
    \begin{tabular}{c|c c c c c c c c|c|c|c|c|c|c}
    \toprule[1pt]
    \hline
    & \multicolumn{8}{c|}{Components} & \multicolumn{3}{c|}{Visible to Thermal} & \multicolumn{3}{c}{Thermal to Visible} \\
    \cline{2-15}
    \multirow{-2}{*}{Order} & $\mathcal{L}_{VCL}$ & $\mathcal{L}_{ICL}$ & $\mathcal{L}_{Tri}$ & $\mathcal{L}_{D}$ & $\mathcal{L}_{CCL}$ & $\mathcal{L}_{MCL}$ & $\mathcal{L}_{CPAL}$ & ${\rm OTPA}$ & Rank-1 & mAP & mINP & Rank-1 & mAP & mINP \\
    \hline
    1 & $\checkmark$ & $\checkmark$ & $\checkmark$ & - & - & - & - & - & 27.65 & 26.45 & 18.54 & 28.98 & 27.88 & 19.22 \\
    2 & $\checkmark$ & $\checkmark$ & $\checkmark^{\star}$ & - & - & - & - & $\checkmark$ & 35.53 & 30.76 & 18.30 & 34.27 & 28.87 & 19.01 \\
    3 & $\checkmark$ & $\checkmark$ & $\checkmark^{\star}$ & $\checkmark$ & - & - & - & $\checkmark$ & 43.07 & 37.14 & 21.98 & 43.11 & 36.35 & 22.08 \\
    4 & $\checkmark$ & $\checkmark$ & $\checkmark^{\star}$ & $\checkmark$ & $\checkmark$ & - & - & $\checkmark$ & 62.23 & 42.97 & 29.81 & 62.34 & 42.53 & 28.98 \\
    5 & $\checkmark$ & $\checkmark$ & $\checkmark^{\star}$ & $\checkmark$ & $\checkmark$ & $\checkmark$ & - & $\checkmark$ & 85.92 & 74.52 & 56.30 & 84.37 & 73.08 & 53.06 \\
    6 & $\checkmark$ & $\checkmark$ & $\checkmark^{\star}$ & $\checkmark$ & $\checkmark$ & $\checkmark$ & $\checkmark$ & $\checkmark$ & 90.34 & 80.72 & 64.29 & 89.04 & 79.50 & 61.27 \\
    \hline
    \bottomrule[1pt]
	\end{tabular}
    }
	\label{Tab:ablation_RegDB}
\end{table*}

\subsubsection{Effectiveness of OTPA Algorithm} 
Apparently, benefiting from OTPA, we achieve a significant promotion of final performance on both datasets (around 25\% Rank-1 (60\% Rank-1) and 20\% mAP (55\% mAP) on SYSU-MM01 (RegDB), by comparing 1$^{\rm st}$, 6$^{\rm th}$ in Tab. \ref{Tab:ablation_SYSU-MM01} (Tab. \ref{Tab:ablation_RegDB}). It appears that OTPA allows us to dig up the cross-modality correspondence and thereby enables us to mine the extra cross-modality supervised signals. It is also worth noting that when using different training sampling strategies, the baseline performance with OTPA (see 2$^{\rm nd}$ in Tab. \ref{Tab:ablation_SYSU-MM01} and Tab. \ref{Tab:ablation_RegDB}) is lower than the pretraining model (see 1$^{\rm st}$ in Tab. \ref{Tab:ablation_SYSU-MM01} and Tab. \ref{Tab:ablation_RegDB}). The difference is caused by the sampling process, where $\mathcal{L}_{VCL}$ and $\mathcal{L}_{ICL}$ are more consistent with the clustering-aware pseudo labels.

\subsubsection{Effectiveness of Contrastive Learning}
Here, we analyze the effectiveness of prototype-based contrastive learning losses. First of all, the pretraining model (1$^{\rm st}$ in Tab. \ref{Tab:ablation_SYSU-MM01} and Tab. \ref{Tab:ablation_RegDB}) and the OTPA baseline (2$^{\rm nd}$ in Tab. \ref{Tab:ablation_SYSU-MM01} and Tab. \ref{Tab:ablation_RegDB}) both achieve relatively good results with the assistance of $\mathcal{L}_{VCL}$ and $\mathcal{L}_{ICL}$, which are comparable to the overall performance of OTLA-ReID as shown in Tab. \ref{Tab:SYSU-MM01} and Tab. \ref{Tab:RegDB}. We can also find that cross contrastive loss ($\mathcal{L}_{CCL}$) improves the performance of around 12\% mAP (15\% mAP) and 15\% Rank-1 (20\% Rank-1) on SYSU-MM01 (RegDB) (see 3$^{\rm rd}$, 4$^{\rm th}$ in Tab. \ref{Tab:ablation_SYSU-MM01} and Tab. \ref{Tab:ablation_RegDB}). The significant gain verifies the effectiveness of cross-modality correspondence generated by OTPA. Besides, the mutual contrastive learning ($\mathcal{L}_{MCL}$) encourages exploring the complementary knowledge between the clustering pseudo labels and cross-modality pseudo labels, which can achieve great improvements (see 4$^{\rm th}$, 5$^{\rm th}$ in Tab. \ref{Tab:ablation_SYSU-MM01} and Tab. \ref{Tab:ablation_RegDB}) as well.

\subsubsection{Effectiveness of Cross Prediction Alignment Learning}
Cross prediction alignment learning can boost performance when combined with other components (see 5$^{\rm th}$, 6$^{\rm th}$ in Tab. \ref{Tab:ablation_SYSU-MM01} and Tab. \ref{Tab:ablation_RegDB}, especially on RegDB). That indicates a further promotion would be expected when aligning the batch-level prototype-based predictions with the mix-up technique between two modalities. However, compared with OTLA-ReID, the improvement in model performance is not very pronounced. One potential explanation is the variance between the prototype-based predictions (cosine similarity based on prototypes) and the classifier's predictions. More detailed analysis will be presented in the discussion section.

\subsubsection{Effectiveness of Critical Tricks}
By reviewing recent works \cite{PGM,MBCCM,DOTLA}, we find the backbone with the training or testing tricks can significantly improve the performance. Tab. \ref{Tab:ablation_tricks} has shown some detailed ablation results of these tricks. From the table, 'Non-Local' denotes whether to use the non-local based ResNet-50 model structure, 'CA' denotes whether to use channel augmentation for visible data, 'Pretrain' denotes whether to use the pretraining strategy , 'LS' denotes whether to use the label smoothing for contrastive learning losses and 'Hflip' denotes whether to use the horizontal flipping data augmentation during the testing stage. Among these tricks, 'Non-Local', 'CA', and 'Pretrain' are shown in the code repository of \cite{PGM} but not in the paper. While 'LS' is a common trick for cross-entropy loss, here we use it for contrastive learning loss.

\renewcommand\arraystretch{1.5}
\begin{table}[h]
	\centering
	\caption{Ablation study of critical tricks in terms of CMC(\%), mAP(\%), and mINP(\%) on SYSU-MM01 (All Search) and RegDB (Visible to Thermal).}
	\resizebox{1.0\columnwidth}{!}{
    \begin{tabular}{c|c c c c c|c|c|c|c|c|c}
    \toprule[1.5pt]
    \hline
    & \multicolumn{5}{c|}{Components} & \multicolumn{3}{c|}{SYSU-MM01 (All Search)} & \multicolumn{3}{c}{RegDB (Visible to Thermal)} \\
    \cline{2-12}
    \multirow{-2}{*}{Order} & Non-Local & CA & Pretrain & LS & Hflip & Rank-1 & mAP & mINP & Rank-1 & mAP & mINP \\
    \hline
    1 & - & - & - & - & - & 49.62 & 44.46 & 29.82 & 77.05 & 71.25 & 57.52 \\
    2 & $\checkmark$ & - & - & - & - & 49.65 & 44.51 & 29.47 & 76.99 & 71.61 & 57.22 \\
    3 & $\checkmark$ & $\checkmark$ & - & - & - & 49.91 & 45.19 & 29.91 & 77.19 & 71.49 & 58.50 \\
    4 & $\checkmark$ & $\checkmark$ & $\checkmark$ & - & - & 53.90 & 51.12 & 35.74 & 83.69 & 75.52 & 63.87 \\
    5 & $\checkmark$ & $\checkmark$ & $\checkmark$ & $\checkmark$ & - & 56.34 & 52.71 & 37.21 & 86.75 & 76.14 & 64.02 \\
    6 & $\checkmark$ & $\checkmark$ & $\checkmark$ & $\checkmark$ & $\checkmark$ & 60.58 & 56.07 & 40.25 & 90.34 & 80.72 & 64.29 \\
    \hline
    \bottomrule[1.5pt]
	\end{tabular}
    }
	\label{Tab:ablation_tricks}
\end{table}

\subsection{Discussion}
\subsubsection{Training Time Analysis}
\begin{figure}[!t]
	\centering
	\includegraphics[scale=0.35]{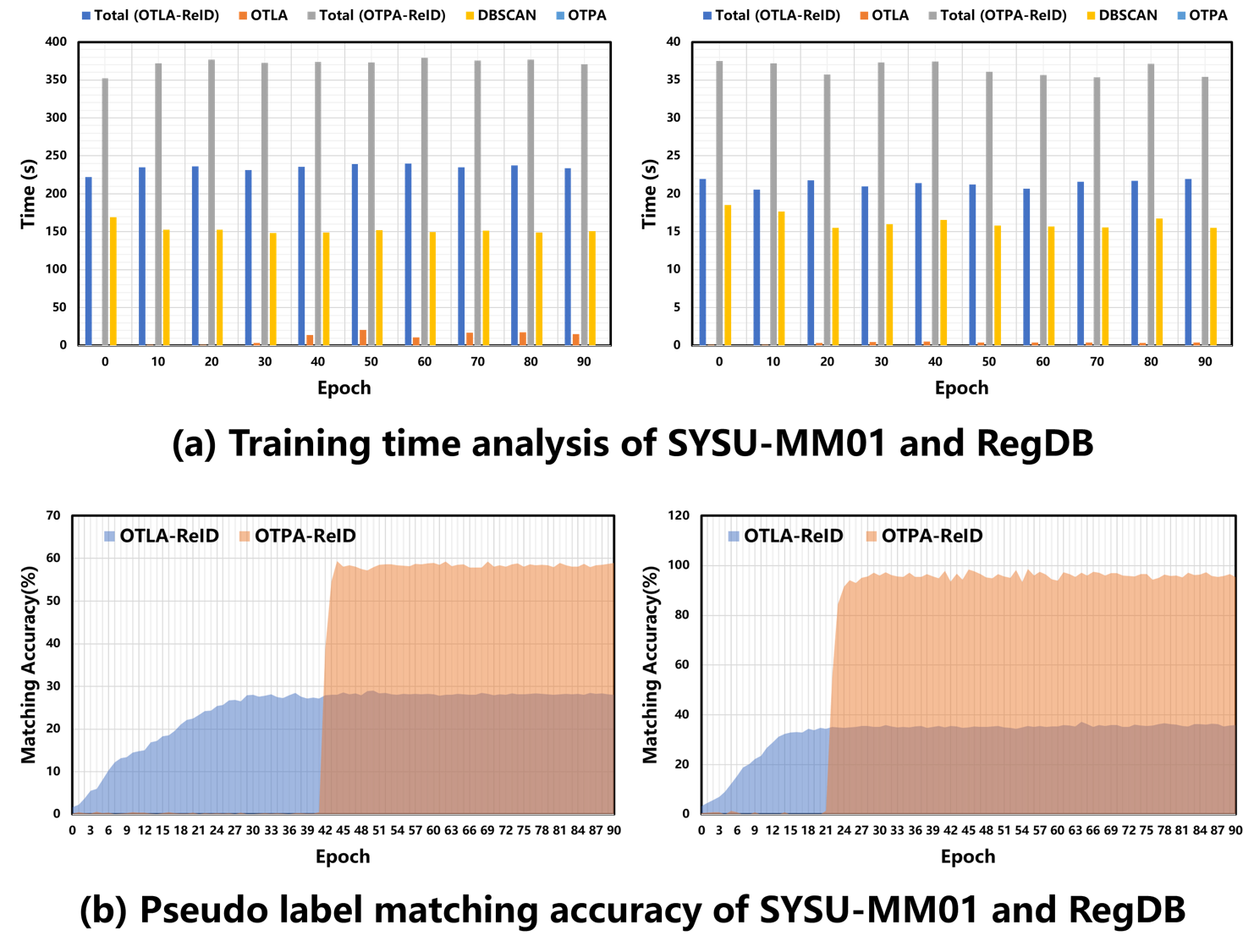}
	\caption{(a) Quantitative analysis of training time. We show the total training time of OTLA-ReID, OTLA running time (the key component in OTLA-ReID \cite{OTLA-ReID} which are most related to OTPA), total training time of our proposed OTPA-ReID, DBSCAN running time, and OTPA running time every 10 training epochs. (b) The comparison of pseudo label matching accuracy for OTLA-ReID \cite{OTLA-ReID} and OTPA-ReID in each training epoch.}
    \label{Fig:trainingtime_matchinglabelaccuracy}
\end{figure}

As illustrated in the Fig. \ref{Fig:trainingtime_matchinglabelaccuracy}(a), we show the total training time of OTLA-ReID, OTLA running time (the key component in OTLA-ReID \cite{OTLA-ReID} which are most related to OTPA), total training time of our proposed OTPA-ReID, DBSCAN running time, and OTPA running time every 10 training epochs. Note that we pretrain the model of OTPA-ReID with 40 (20) epochs on SYSU-MM01 (RegDB) so the OTPA running time of the pretraining epoch is zero. Apparently, we can see OTPA algorithm is very efficient for both datasets as it only transports the prototypes that are relatively fewer in number, where the running time is negligible (around 0.01s for SYSU-MM01 and 0.005s for RegDB) compared with OTLA algorithm. However, OTPA algorithm introduces the extra DBSCAN for preliminary clustering (around the 150s for SYSU-MM01 and 15s for RegDB) due to the lack of annotation. Actually, OTLA algorithm also necessitates an additional UDA process, which is more time-consuming and resource-intensive, to generate reliable visible pseudo labels in the unsupervised setting.

\begin{figure*}[!t]
	\centering
	\includegraphics[scale=0.165]{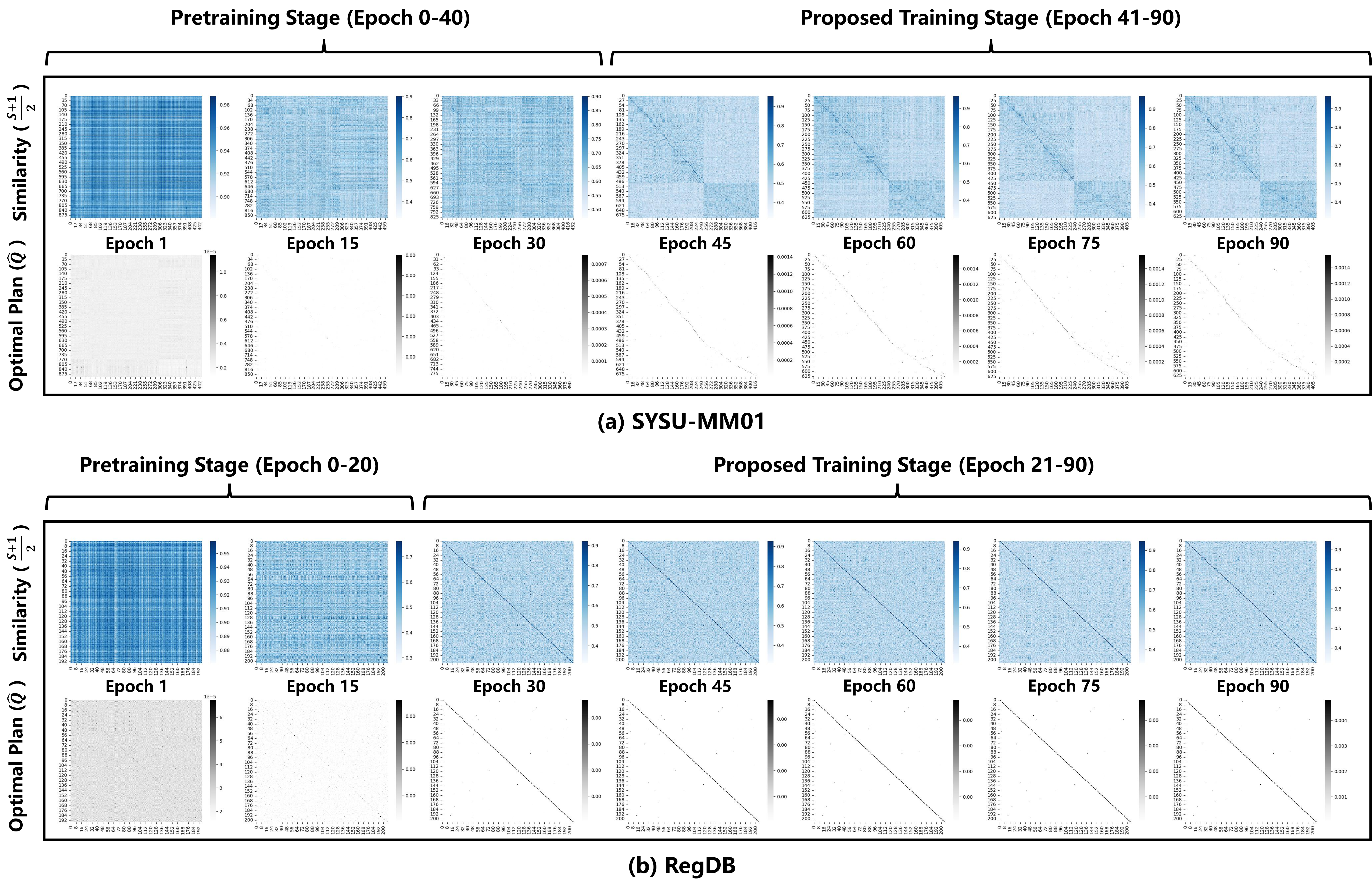}
	\caption{Visualization of optimal transport prototype assignment (OTPA) algorithm. We visualize the normalized cosine similarity matrix $\frac{\boldsymbol{S}+1}{2}$ (input of OTPA algorithm ranged from $[0,1]$) and the optimal transport plan $\hat{\boldsymbol{Q}}$ (output of OTPA algorithm). The horizontal and vertical coordinates of each visualized matrix denote visible pseudo classes and infrared pseudo classes, respectively.}
	\label{Fig:OTPAvisualization}
\end{figure*}

% \begin{figure}[!t]
% 	\centering
% 	\includegraphics[scale=0.265]{image/OTPAquantitativeanalysis.png}
% 	\caption{Quantitative analysis of optimal-transport prototype assignment w/wo uniform prior on SYSU-MM01 (All Search) and RegDB (Visible to Infrared) datasets.}
% 	\label{Fig:OTPAquantitativeanalysis}
% \end{figure}

\subsubsection{Pseudo Labels Matching Accuracy}
To illustrate the accuracy of our generated pseudo labels, we show the matching accuracy of cross-modality pseudo labels generated by OTPA and OTLA (unsupervised) in Fig. \ref{Fig:trainingtime_matchinglabelaccuracy}(b). This epoch-level matching accuracy is defined by the average proportion of our learned cross-modal correspondence and the ground-truth for the cross-modality data pairs. The cross-modality data pairs are constructed by randomly selecting two samples sharing the same cross-modality pseudo label (Note that both OTLA and OTPA can produce the matching data pair). Due to the initial 40 (20) training epochs of OTPA-ReID dedicated to pretraining on SYSU-MM01 (RegDB), only cluster-aware pseudo labels are available for sampling training data. Therefore, the corresponding matching accuracy is nearly zero. The visualization results show that the pseudo label matching accuracy of OTLA-ReID and OTPA-ReID iteratively improved as the training goes on. However, OTPA-ReID can finally reach around 58\% on SYSU-MM01 and around 95\% on RegDB significantly outperforming OTLA-ReID. This phenomenon indicates that the OTPA algorithm can more effectively explore the correct cross-modality correspondence. Besides, this pseudo label matching accuracy is very close to the Rank-1 accuracy, which can be an important indicator for the USVI-ReID task.

\subsubsection{Visualization of Feature Embedding Space}
\begin{figure}[!t]
	\centering
	\includegraphics[scale=0.55]{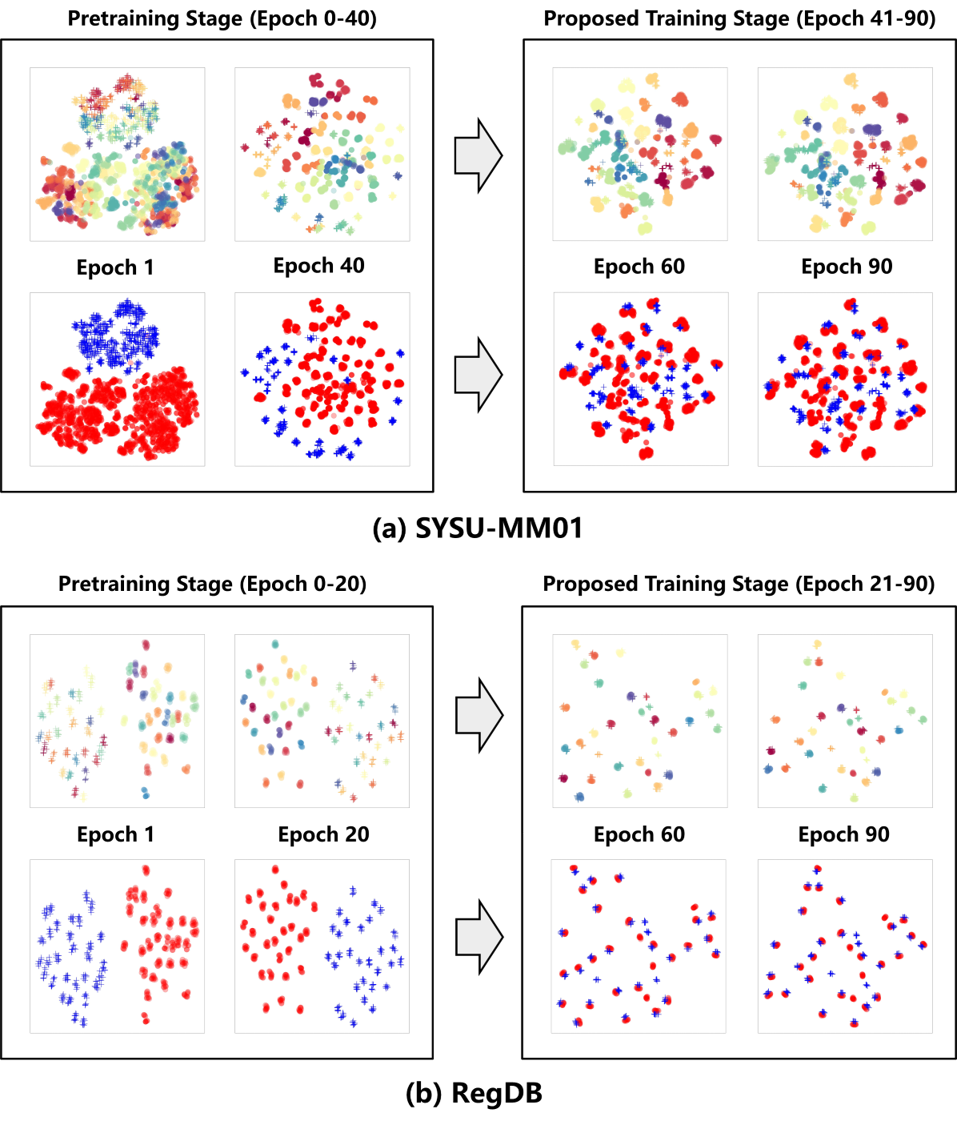}
	\caption{Visualization of feature embedding space with t-SNE \cite{t-SNE} on SYSU-MM01 and RegDB, where we randomly select cross-modality samples from 20 classes. The circle symbol ($\circ$) denotes the instance from visible modality and the cross symbol ($+$) denotes the instance from infrared modality. The different colors of the upper figure of each dataset indicate the samples from different classes, while in the bottom figure, two kinds of colors only reflect the modality information.}
	\label{Fig:EmbeddingSpaceTSNE}
\end{figure}

We visualize the feature embedding space with t-SNE \cite{t-SNE} by projecting the features into a 2D plane. We randomly select the cross-modality samples from 20 classes in Fig. \ref{Fig:EmbeddingSpaceTSNE}. It appears that at the pretraining stage (\textit{i.e.}, epoch 0-40 for SYSU-MM01 and epoch 0-20 for RegDB), the inter-modality discrepancy is large so that there are obvious intervals between the cross-modality sample points. We can also find that sample points of different classes within one modality can not be separated explicitly in this initial stage. However, when incorporating our proposed method, the points with the same ground-truth labels are gradually gathering together as the training epoch increases no matter what modality they come from. This phenomenon vividly validates the effectiveness of our proposed method, although there still exists wrongly assembled sample points.

\begin{figure*}[!t]
	\centering
	\includegraphics[scale=0.54]{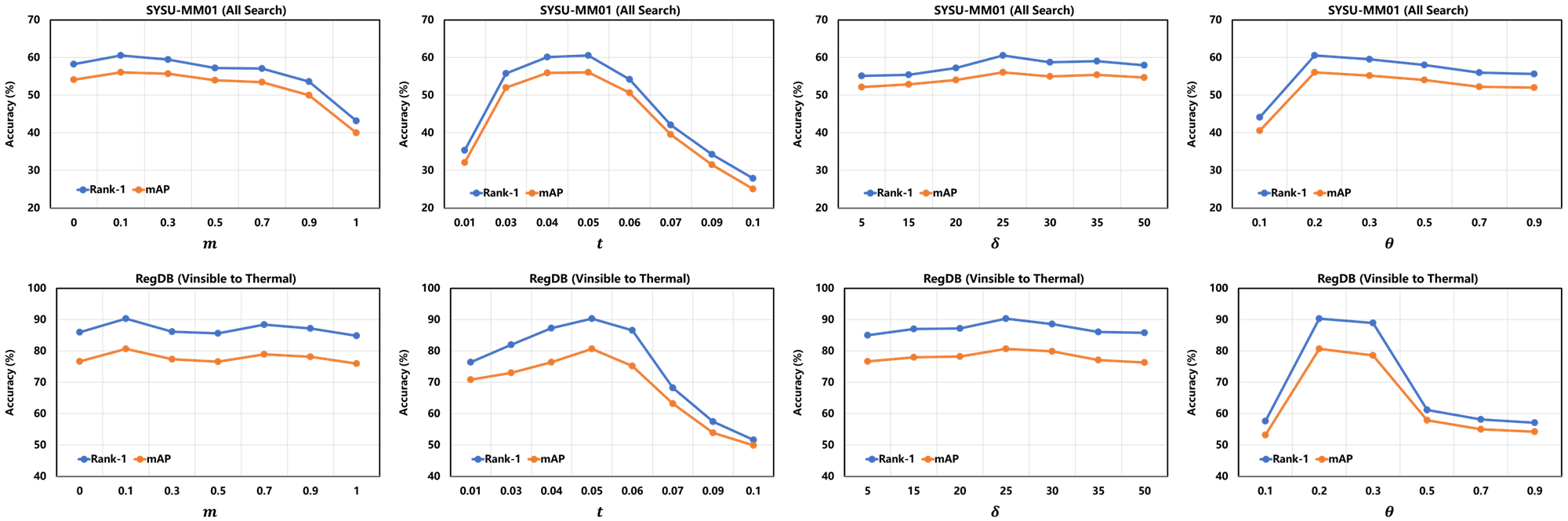}
	\caption{Sensitivity experiments corresponding to several critical hyper-parameters on both SYSU-MM01 and RegDB datasets. Each line of the figure is the corresponding sensitive experiment belonging to one of the hyper-parameters.}
	\label{Fig:sensitivityanalysis}
\end{figure*}

% \begin{figure*}[!t]
% 	\centering
% 	\includegraphics[scale=0.54]{image/CPALvisualization.png}
% 	\caption{\textcolor{red}{Visualization of cross prediction alignment learning (CPAL) on SYSU-MM01 and RegDB. The upper figures are the batch-level ground-truth binary matrices indicating whether the data pairs are matched. The bottom figures are our learned attention coefficient matrices of CPAL.}}
% 	\label{Fig:CPALvisualization}
% \end{figure*}

\subsubsection{Visualization of OTPA Algorithm}
As shown in Fig. \ref{Fig:OTPAvisualization}(a) and Fig. \ref{Fig:OTPAvisualization}(b), we visualize the optimal transport results. The upper parts of this figure depict the normalized cosine similarity matrices $\frac{\boldsymbol{S}+1}{2}$ (input of OTPA algorithm ranged from $[0,1]$) computed between the prototypes from two modality-specific memories, while the bottom parts show the optimal transport plan $\hat{\boldsymbol{Q}}$ (output of OTPA algorithm). Darker colors indicate higher values. The horizontal and vertical coordinates of each visualized matrix denote the pseudo classes for visible and infrared prototype memory, respectively. Note that we do not use OTPA algorithm for pretraining so we implement it here only for better visualization and comparison.

As observed, irrespective of the utilization of the OTPA algorithm, we are unable to establish clear cross-modality correspondence during the pretraining stage. This is because the pretraining sampling strategy we used relies on intra-modality clustering results. It also can be seen that as our proposed training stage progresses, the cross-modality correspondence becomes more and more explicit, as it is optimized by our designed OTPA algorithm and contrastive learning losses. Furthermore, it seems that OPTA algorithm can generate more explicit correspondence compared with the cosine similarity matrix ({\it i.e.,} without OTPA algorithm). Besides, with the uniform prior ({\it i.e.,}$\boldsymbol{\alpha}$ and $\boldsymbol{\beta}$) in OTPA, the degeneration phenomenon (\textit{i.e.}, most prototypes of one modality are assigned to few prototypes from another modality) will be relieved in the transporting process.

\subsubsection{Sensitivity Analysis}
For this part, we would like to give some sensitivity analyses of several critical hyper-parameters on both SYSU-MM01 and RegDB datasets, which are shown in Fig. \ref{Fig:sensitivityanalysis}.

\textbf{Memory Momentum Value ($\boldsymbol{m}$).} We use the momentum updating strategy to renew both modality-specific prototype memories and cross prototype memory per training batch. The memory momentum value $m$ in Eq. (\ref{memory update 1}-\ref{memory update 3}) controls how much new knowledge of each training batch's data will be merged into these prototype memories. In other words, the higher $m$ value means more integration of the current batch's data. From Fig. \ref{Fig:sensitivityanalysis}, we can find the model achieves the best results when $m$ equals 0.1 on both datasets. Besides, the memory momentum value seems more robust to RegDB than SYSU-MM01.

\textbf{Temperature ($\boldsymbol{\tau}$).} The temperature $\boldsymbol{\tau}$ is used to adjust the softness of the cosine similarity score when computing prototype-based contrastive learning loss and cross prediction alignment learning loss. It seems to be a highly sensitive hyper-parameter as shown in Fig. \ref{Fig:sensitivityanalysis}. 
Obviously, the temperature coefficient of 0.05 is the most appropriate value for our model.

\textbf{Smoothness of OTPA ($\boldsymbol{\delta}$).} The hyper-parameter $\delta$ is used to determine the smoothness of OTPA, which trades off convergence speed with closeness to the optimal transport problem. This hyper-parameter is quite stable with the performance. According to Fig. \ref{Fig:sensitivityanalysis}, we choose the value of 25 for ${\delta}$, which is slightly superior to other values.

\textbf{Threshold of $\boldsymbol{\mathcal{L}_{MCL}}$ ($\boldsymbol{\theta}$).} We also analyze the sensitivity of hyper-parameter ${\theta}$, which is a threshold to filter the unreliable similarity scores when computing mutual contrastive learning loss. The larger value of ${\theta}$ means that fewer positive similarity scores participated in the calculation. As shown in Fig. \ref{Fig:sensitivityanalysis}, this hyper-parameter seems more robust to SYSU-MM01 than RegDB. For both datasets, the value of 0.2 can be the best choice for our model.

% \textbf{Coefficients of $\boldsymbol{\mathcal{L}_{CPAL}}$ ($\boldsymbol{\mu_{1}}, \boldsymbol{\mu_{2}}$).} These two hyper-parameters decide the weight (importance) of the corresponding part for cross prediction alignment learning loss. We fix one of the hyper-parameters and change the other one to show its influence. Surprisingly, they are more robust than other hyper-parameters referring to Fig. \ref{Fig:sensitivityanalysis}. Besides, 0.5 for $\mu_{1}$ and 1.0 for $\mu_{2}$ seem to be proper weights for this loss.

\renewcommand\arraystretch{1.1}
\begin{table*}[h]
	\centering
	\caption{The performance comparison between the pertaining stage (commonly used in ADCA \cite{ADCA} and PGM \cite{PGM}) and our proposed training stage with different $\gamma$ values on SYSU-MM01 (All Search) and RegDB (Visible to Thermal). $\gamma$ is used to control the proportion of the overlapping classes between the visible subset and the infrared subset. We propose the Share mode and Share+Indep mode to verify the robustness of our method. All experiments are measured by CMC(\%), mAP(\%) and mINP(\%).}
	\resizebox{2.0\columnwidth}{!}{
		\begin{tabular}{c|c|c|c|c|c|c|c|c}
            \toprule[1pt]
			\hline
			\multicolumn{3}{c|}{Settings} &
			\multicolumn{3}{c|}{SYSU-MM01 (All Search)} &
			\multicolumn{3}{c}{RegDB (Visible to Thermal)} \\
			\hline
			Mode & $\gamma$ value & Stage & Rank-1 & mAP & mINP & Rank-1 & mAP & mINP \\
			\hline
			\multirow{12}{*}{Share + Indep}
			& \multirow{2}{*}{0.0} & Pretrain & 24.50 & 24.59 & 13.90 & 10.53 & 12.20 & 7.97 \\
            & & Train & 23.56 (\textcolor{green}{-0.94}) & 23.59 (\textcolor{green}{-1.00}) & 12.83 (\textcolor{green}{-1.07}) & 26.02 (\textcolor{red}{+15.49}) & 22.83 (\textcolor{red}{+10.63}) & 13.70 (\textcolor{red}{+5.73}) \\
            \cline{2-9}
            & \multirow{2}{*}{0.1} & Pretrain & 26.93 & 26.47 & 14.70 & 12.23 & 12.88 & 8.15 \\
            & & Train & 29.57 (\textcolor{red}{+2.64}) & 28.90 (\textcolor{red}{+2.53}) & 16.86 (\textcolor{red}{+2.16}) & 34.27 (\textcolor{red}{+22.04}) & 31.06 (\textcolor{red}{+18.18}) & 19.31 (\textcolor{red}{+11.16}) \\
            \cline{2-9}
            & \multirow{2}{*}{0.3} & Pretrain & 29.39 & 28.36 & 15.82 & 14.61 & 15.21 & 8.44 \\
            & & Train & 35.63 (\textcolor{red}{+6.24}) & 33.50 (\textcolor{red}{+5.14}) & 19.94 (\textcolor{red}{+4.12}) & 41.84 (\textcolor{red}{+27.23}) & 37.53 (\textcolor{red}{+22.32}) & 24.24 (\textcolor{red}{+15.79}) \\
            \cline{2-9}
            & \multirow{2}{*}{0.5} & Pretrain & 34.98 & 33.81 & 20.71 & 20.29 & 20.21 & 12.13 \\
            & & Train & 48.80 (\textcolor{red}{+13.82}) & 45.64 (\textcolor{red}{+11.83}) & 31.01 (\textcolor{red}{+10.3}) & 70.58 (\textcolor{red}{+50.29}) & 61.25 (\textcolor{red}{+41.04}) & 42.48 (\textcolor{red}{+30.35}) \\
            \cline{2-9}
			& \multirow{2}{*}{0.7} & Pretrain & 34.16 & 32.85 & 19.68 & 23.45 & 21.91 & 13.25 \\
            & & Train & 51.11 (\textcolor{red}{+16.95}) & 47.75 (\textcolor{red}{+14.90}) & 32.22 (\textcolor{red}{+12.54}) & 81.70 (\textcolor{red}{+58.25}) & 70.39 (\textcolor{red}{+48.48}) & 52.21 (\textcolor{red}{+38.96}) \\
            \cline{2-9}
            & \multirow{2}{*}{0.9} & Pretrain & 34.77 & 33.75 & 21.06 & 26.45 & 25.69 & 17.45 \\
            & & Train & 56.35 (\textcolor{red}{+21.58}) & 52.32 (\textcolor{red}{+18.57}) & 36.53 (\textcolor{red}{+15.47}) & 87.33 (\textcolor{red}{+60.88}) & 77.41 (\textcolor{red}{+51.72}) & 61.02 (\textcolor{red}{+43.57}) \\
            \hline
			\multirow{12}{*}{Share} & \multirow{2}{*}{0.1} & Pretrain & 5.37 & 6.97 & 2.84 & 4.37 & 5.67 & 3.66 \\
            & & Train & 6.81 (\textcolor{red}{+1.44}) & 7.62 (\textcolor{red}{+0.65}) & 2.92 (\textcolor{red}{+0.08}) & 21.17 (\textcolor{red}{+16.8}) & 19.57 (\textcolor{red}{+13.9}) & 10.20 (\textcolor{red}{+6.54}) \\
            \cline{2-9}
            & \multirow{2}{*}{0.3} & Pretrain & 15.23 & 16.00 & 7.74 & 6.99 & 7.39 & 4.52 \\
            & & Train & 20.71 (\textcolor{red}{+5.48}) & 20.16 (\textcolor{red}{+4.16}) & 9.72 (\textcolor{red}{+1.98}) & 44.71 (\textcolor{red}{+37.72}) & 36.41 (\textcolor{red}{+29.02}) & 20.05 (\textcolor{red}{+15.53}) \\
            \cline{2-9}
            & \multirow{2}{*}{0.5} & Pretrain & 23.04 & 23.03 & 12.47 & 5.83 & 8.38 & 5.07 \\
            & & Train & 36.97 (\textcolor{red}{+13.93}) & 34.96 (\textcolor{red}{+11.93}) & 21.17 (\textcolor{red}{+8.7}) & 68.06 (\textcolor{red}{+62.23}) & 57.02 (\textcolor{red}{+48.64}) & 38.46 (\textcolor{red}{+33.39}) \\
            \cline{2-9}
            & \multirow{2}{*}{0.7} & Pretrain & 32.04 & 30.70 & 17.89 & 15.49 & 15.90 & 9.76 \\
            & & Train & 49.99 (\textcolor{red}{+17.95}) & 45.53 (\textcolor{red}{+14.83}) & 29.84 (\textcolor{red}{+11.95}) & 83.06 (\textcolor{red}{+67.57}) & 70.86 (\textcolor{red}{+54.96}) & 52.40 (\textcolor{red}{+42.64}) \\
            \cline{2-9}
            & \multirow{2}{*}{0.9} & Pretrain & 33.46 & 32.38 & 19.29 & 21.12 & 21.27 & 12.75 \\
            & & Train & 56.47 (\textcolor{red}{+23.01}) & 51.91 (\textcolor{red}{+19.53}) & 36.03 (\textcolor{red}{+16.74}) & 89.37 (\textcolor{red}{+68.25}) & 78.47 (\textcolor{red}{+57.23}) & 60.48 (\textcolor{red}{+48.03}) \\
            \cline{2-9}
            & \multirow{2}{*}{1.0} & Pretrain & 36.30 & 35.41 & 22.14 & 27.65 & 26.45 & 18.54 \\
            & & Train & 60.58 (\textcolor{red}{+24.28}) & 56.07 (\textcolor{red}{+20.66}) & 40.25 (\textcolor{red}{+18.11}) & 90.34 (\textcolor{red}{+62.69}) & 80.72 (\textcolor{red}{+54.27}) & 64.29 (\textcolor{red}{+45.75}) \\
			\hline
            \bottomrule[1pt]
		\end{tabular}
	}
	\label{Tab:different_class_proportion}
\end{table*}

\subsubsection{Analysis of Inconsistent Class Space in Training Data}
As we can see, the majority of recent works \cite{H2H,ADCA,OTLA-ReID,CHCR,DOTLA,MBCCM,CCLNet,PGM} on the unsupervised VI-ReID task (including ours) are based on a strong assumption that the training data are exactly the same set of pedestrians in different modalities. To explore the influence of this assumption, we have done extra experiments (\textit{i.e.}, we select a subset of the original training set, and then set a hyper-parameter $\gamma$ to control the proportion of the overlapped classes between the visible subset and the infrared subset. The remaining classes are randomly divided with equal proportion for each modality).

\begin{figure}[!t]
	\centering
	\includegraphics[scale=0.26]{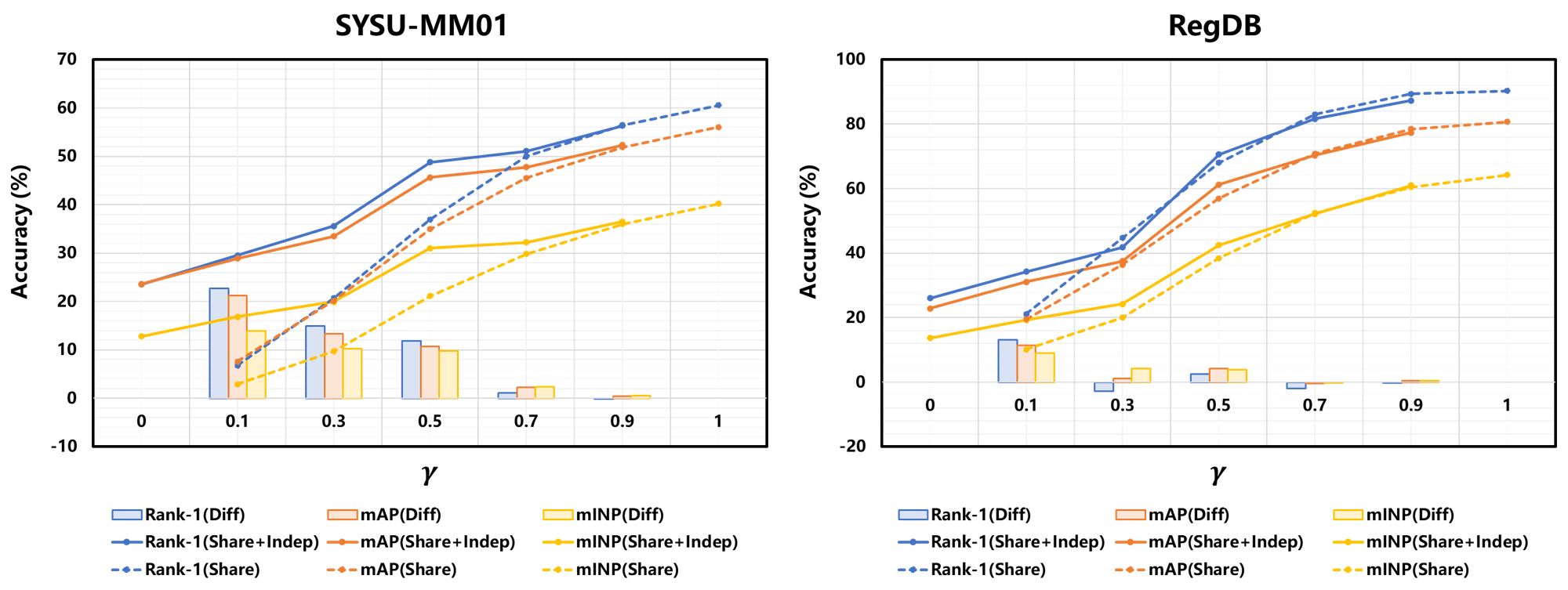}
	\caption{Performance under the Share mode and the Share+Indep mode as the $\gamma$ value changing. We also visualize the performance's difference (Diff, \textit{i.e.}, Share+Indep mode minus Share mode) of the same $\gamma$ value.} 
	\label{Fig:different_class_proportion}
\end{figure}

For example, suppose there are 4 categories (\textit{i.e.}, 1, 2, 3, 4) in the training set. If we set $\gamma =0.5$, then we randomly select 2, 3 as the overlapped classes. Share mode indicates we only take the cross-modality samples from 2, 3 for training. Share+Indep mode indicates that the visible training set contains categories 1, 2, 3 while the infrared set contains categories 2, 3, 4. For better comparison, we also do the experiments of Share mode by only using the category overlapped cross-modality training data. Specifically, we report the average experimental results in Tab. \ref{Tab:different_class_proportion} with $\gamma$ ranging from $0$ to $1$. Note that each mode here is conducted by repeating the experiment 3 times to eliminate the effect of the random seed. Besides, we also visualize the model's performance under the Share mode and the Share+Indep mode as the $\gamma$ value changing in Fig. \ref{Fig:different_class_proportion}. We also add the performance's difference (Diff, \textit{i.e.}, Share+Indep mode minus Share mode) of the same $\gamma$ value into the Fig. \ref{Fig:different_class_proportion}. 

From Tab. \ref{Tab:different_class_proportion} and Fig. \ref{Fig:different_class_proportion}, we can find that our proposed method is robust and consistently improves the performances compared with the pretraining stage (note that the pretraining and our cross-modality training use the same data), even the the classes of each modality are incompletely non-overlapping. When $\gamma$ is below than 0.9,  Share+Indep mode exceeds the Share mode. When the classes are nearly overlapping ($\gamma = 0.9$), it appears that these two modes achieve comparable performance.

In conclusion, we can see this assumption is necessary and important in the unsupervised cross-modality person ReID task. Surprisingly, even when the classes of each modality are incompletely overlapping, our approach still works. This is because identity-aware discrimination does not rely solely on fully supervised signals. For example, people dressed similarly also can provide useful information. On the other hand, it is usual to collect data from a supermarket or mall during a period of time. Thus one person appearing in multiple cameras is quite common. 

\section{Conclusion}
In this paper, we propose a mutual information guided optimal transport approach for unsupervised VI-ReID. In this framework, three learning principles, \textit{i.e.}, "Sharpness" (entropy minimization), "Fairness" (uniform label distribution) and "Fitness" (cross-modality matching) are derived based on the proposed theoretical objective. To reach these principles, a loop iterative training strategy alternating between model training and cross-modality matching is designed. In the matching stage, we formulate the cross-modality prototype matching problem as an optimal transport task, which allows for finding the cross-modality correspondence. In the training stage, prototype-based contrastive learning losses are designed for cross-modality learning based on this correspondence. In future work, we would like to do more theoretical analysis and propose more generalized methods for different unsupervised cross-modality learning tasks, \textit{e.g.}, unsupervised vision-language learning task.

\appendices
\section{Proof Details}
% \subsection{Properties of Mutual Information}
% In this section, we give some properties of mutual information used in our proposed theoretical proof. 

\subsection{Derivation from Eq. (3) to Eq. (5)}
\noindent\textbf{Theorem 1.} \textit{For random variables $\boldsymbol{x}^{1},\boldsymbol{x}^{2},\boldsymbol{y}^{1},\boldsymbol{y}^{2}$, we can observe the following correlation:
\begin{equation}
\begin{aligned}
    &\max\ I(\boldsymbol{x}^{1};\boldsymbol{y}^{1}) + I(\boldsymbol{x}^{2};\boldsymbol{y}^{1}) + I(\boldsymbol{x}^{1};\boldsymbol{y}^{2}) + I(\boldsymbol{x}^{2};\boldsymbol{y}^{2}) \\ &\Leftrightarrow \max\ I(\boldsymbol{y}^{1};\boldsymbol{x}^{1},\boldsymbol{x}^{2}) + I(\boldsymbol{y}^{2};\boldsymbol{x}^{1},\boldsymbol{x}^{2}),
    \label{a-1}
\end{aligned}
\end{equation}
where $\boldsymbol{y}^{1},\boldsymbol{y}^2$ are dependent on the  $\boldsymbol{x}^{1},\boldsymbol{x}^{2}$, respectively.}

\noindent\textbf{Proof of Theorem 1.} We can reformulate the left part of Eq. (\ref{a-1}) according to the ($P_{3}$) chain rule (Multivariate Mutual Information) of \cite{MIB} in appendix, \textit{i.e.}, for random variables $\boldsymbol{x},\boldsymbol{y},\boldsymbol{z}$, then $I(\boldsymbol{y};\boldsymbol{z}) = I(\boldsymbol{y};\boldsymbol{z};\boldsymbol{x}) + I(\boldsymbol{y};\boldsymbol{z}|\boldsymbol{x})$. The equivalent equation is:
\begin{equation}
\begin{aligned}
    \max\ &\underbrace{2I(\boldsymbol{y}^{1};\boldsymbol{x}^{1};\boldsymbol{x}^{2}) + I(\boldsymbol{y}^{1};\boldsymbol{x}^{1}|\boldsymbol{x}^{2}) + I(\boldsymbol{y}^{1};\boldsymbol{x}^{2}|\boldsymbol{x}^{1})}_{I(\boldsymbol{y}^{1};\boldsymbol{x}^{1}) 
    + I(\boldsymbol{y}^{2};\boldsymbol{x}^{1})} + \\ &\underbrace{2I(\boldsymbol{y}^{2};\boldsymbol{x}^{2};\boldsymbol{x}^{1}) + I(\boldsymbol{y}^{2};\boldsymbol{x}^{2}|\boldsymbol{x}^{1}) + I(\boldsymbol{y}^{2};\boldsymbol{x}^{1}|\boldsymbol{x}^{2})}_{I(\boldsymbol{y}^{2};\boldsymbol{x}^{2}) + I(\boldsymbol{y}^{2};\boldsymbol{x}^{1})}.
    \label{a-2}
\end{aligned}
\end{equation}

Note that we can remove the constant weights from Eq. (\ref{a-2}), which does not affect the optimization objective. Besides, considering the ($P_{2}$) chain rule of \cite{MIB} in appendix, \textit{i.e.}, for random variables $\boldsymbol{x},\boldsymbol{y},\boldsymbol{z}$, then $I(\boldsymbol{z};\boldsymbol{x},\boldsymbol{y}) = I(\boldsymbol{y};\boldsymbol{z}) + I(\boldsymbol{x};\boldsymbol{z}|\boldsymbol{y})$, the Eq. (\ref{a-2}) can be changed into:
\begin{equation}
\begin{aligned}
    &\max\ \underbrace{I(\boldsymbol{y}^{1};\boldsymbol{x}^{1};\boldsymbol{x}^{2}) + I(\boldsymbol{y}^{1};\boldsymbol{x}^{1}|\boldsymbol{x}^{2}) + I(\boldsymbol{y}^{1};\boldsymbol{x}^{2}|\boldsymbol{x}^{1})}_{\max\ 2I(\boldsymbol{y}^{1};\boldsymbol{x}^{1};\boldsymbol{x}^{2}) \ \Leftrightarrow \ \max\ I(\boldsymbol{y}^{1};\boldsymbol{x}^{1};\boldsymbol{x}^{2})} + \\ &\underbrace{I(\boldsymbol{y}^{2};\boldsymbol{x}^{2};\boldsymbol{x}^{1}) + I(\boldsymbol{y}^{2};\boldsymbol{x}^{2}|\boldsymbol{x}^{1}) + I(\boldsymbol{y}^{2};\boldsymbol{x}^{1}|\boldsymbol{x}^{2})}_{\max\ 2I(\boldsymbol{y}^{2};\boldsymbol{x}^{2};\boldsymbol{x}^{1}) \ \Leftrightarrow \ \max\ I(\boldsymbol{y}^{2};\boldsymbol{x}^{2};\boldsymbol{x}^{1})} \\ &\ \Leftrightarrow \ \max\ \underbrace{I(\boldsymbol{y}^{1};\boldsymbol{x}^{1})}_{I(\boldsymbol{y}^{1};\boldsymbol{x}^{1};\boldsymbol{x}^{2}) + I(\boldsymbol{y}^{1};\boldsymbol{x}^{1}|\boldsymbol{x}^{2})} + I(\boldsymbol{y}^{1};\boldsymbol{x}^{2}|\boldsymbol{x}^{1}) + \\ &\underbrace{I(\boldsymbol{y}^{2};\boldsymbol{x}^{2})}_{I(\boldsymbol{y}^{2};\boldsymbol{x}^{2};\boldsymbol{x}^{1}) + I(\boldsymbol{y}^{2};\boldsymbol{x}^{2}|\boldsymbol{x}^{1})} + I(\boldsymbol{y}^{2};\boldsymbol{x}^{1}|\boldsymbol{x}^{2}) \\ &\ \Leftrightarrow \ \max\ \underbrace{I(\boldsymbol{y}^{1};\boldsymbol{x}^{1},\boldsymbol{x}^{2})}_{I(\boldsymbol{y}^{1};\boldsymbol{x}^{1}) + I(\boldsymbol{y}^{1};\boldsymbol{x}^{2}|\boldsymbol{x}^{1})} + \underbrace{I(\boldsymbol{y}^{2};\boldsymbol{x}^{1},\boldsymbol{x}^{2})}_{I(\boldsymbol{y}^{2};\boldsymbol{x}^{2}) + I(\boldsymbol{y}^{2};\boldsymbol{x}^{1}|\boldsymbol{x}^{2})}.
\label{a-3}
\end{aligned}
\end{equation}

In conclusion, Theorem 1 is proven.

After that, we replace the $\boldsymbol{x}^{1},\boldsymbol{x}^{2}$ with cross-modality variables $\boldsymbol{x}^{v},\boldsymbol{x}^{r}$ and replace $\boldsymbol{y}^{1},\boldsymbol{y}^{2}$ with corresponding predicted label variables $\boldsymbol{y}^{v},\boldsymbol{y}^{r}$. Then we can derive Eq. (5) from Eq. (3) in the main body of the paper:
\begin{equation}
\begin{aligned}
    &\max\ I(\boldsymbol{x}^{v};\boldsymbol{y}^{v}) + I(\boldsymbol{x}^{r};\boldsymbol{y}^{v}) + I(\boldsymbol{x}^{r};\boldsymbol{y}^{r}) + I(\boldsymbol{x}^{v};\boldsymbol{y}^{r}) \\ &\ \Leftrightarrow \ \max\ I(\boldsymbol{y}^{v};\boldsymbol{x}^{v},\boldsymbol{x}^{r}) + I(\boldsymbol{y}^{r};\boldsymbol{x}^{v},\boldsymbol{x}^{r}).
\label{a-3}
\end{aligned}
\end{equation}

\subsection{The Variational Lower Bound of Eq. (7) and Eq. (8)}
% \textcolor{red}{Here, we first propose a theorem, and then give the corresponding proof of it.}

\noindent\textbf{Theorem 2.} \textit{For random variables $\boldsymbol{x},\boldsymbol{y},\boldsymbol{z}$, if $\boldsymbol{z}$ is dependent on the union of $\boldsymbol{x}$ and $\boldsymbol{y}$, \textit{i.e.}, $p(\boldsymbol{z}|\boldsymbol{x},\boldsymbol{y}) > 0$, then we can obtain the variational lower bound $\tilde{I}(\boldsymbol{z};\boldsymbol{x},\boldsymbol{y})$ of $I(\boldsymbol{z};\boldsymbol{x},\boldsymbol{y})$ with KL divergence constraint ${\rm KL}(p(\boldsymbol{z}|\boldsymbol{x},\boldsymbol{y}) || q(\boldsymbol{z}|\boldsymbol{x},\boldsymbol{y})) = \int p(\boldsymbol{z}|\boldsymbol{x},\boldsymbol{y}){\rm log}(\frac{p(\boldsymbol{z}|\boldsymbol{x},\boldsymbol{y})}{q(\boldsymbol{z}|\boldsymbol{x},\boldsymbol{y})}) d\boldsymbol{z} \ge 0$:
\begin{equation}
\begin{aligned}
    I(\boldsymbol{z};\boldsymbol{x},\boldsymbol{y}) &\ge H(\mathbbm{E}_{\boldsymbol{x},\boldsymbol{y}}[q(\boldsymbol{z}|\boldsymbol{x},\boldsymbol{y})]) - \mathbbm{E}_{\boldsymbol{x},\boldsymbol{y}}[H(q(\boldsymbol{z}|\boldsymbol{x},\boldsymbol{y}))] \\ &\overset{{\rm def}}{=} \tilde{I}(\boldsymbol{z};\boldsymbol{x},\boldsymbol{y}),
\label{b-1}
\end{aligned}
\end{equation}
where $q(\boldsymbol{z}|\boldsymbol{x},\boldsymbol{y})$ is an arbitrary variational distribution.}

\noindent\textbf{Proof of Theorem 2.} Let's begin with the definition of $I(\boldsymbol{z};\boldsymbol{x},\boldsymbol{y})$:
\begin{equation}
\begin{aligned}
    &I(\boldsymbol{z};\boldsymbol{x},\boldsymbol{y}) = \iiint p(\boldsymbol{z},\boldsymbol{x},\boldsymbol{y}){\rm log}(\frac{p(\boldsymbol{z},\boldsymbol{x},\boldsymbol{y})}{p(\boldsymbol{x},\boldsymbol{y})p(\boldsymbol{z})})d\boldsymbol{z}d\boldsymbol{x}d\boldsymbol{y} \\ &= \iint p(\boldsymbol{x},\boldsymbol{y})d\boldsymbol{x}d\boldsymbol{y} \int p(\boldsymbol{z}|\boldsymbol{x},\boldsymbol{y}){\rm log}(\frac{p(\boldsymbol{z}|\boldsymbol{x},\boldsymbol{y})}{p(\boldsymbol{z})})d\boldsymbol{z} \\ &= \iint p(\boldsymbol{x},\boldsymbol{y})d\boldsymbol{x}d\boldsymbol{y} \int p(\boldsymbol{z}|\boldsymbol{x},\boldsymbol{y}){\rm log}(\frac{q(\boldsymbol{z}|\boldsymbol{x},\boldsymbol{y})}{p(\boldsymbol{z})}\frac{p(\boldsymbol{z}|\boldsymbol{x},\boldsymbol{y})}{q(\boldsymbol{z}|\boldsymbol{x},\boldsymbol{y})})d\boldsymbol{z} \\ &= \iint p(\boldsymbol{x},\boldsymbol{y})d\boldsymbol{x}d\boldsymbol{y} (\int p(\boldsymbol{z}|\boldsymbol{x},\boldsymbol{y}){\rm log}(\frac{q(\boldsymbol{z}|\boldsymbol{x},\boldsymbol{y})}{p(\boldsymbol{z})})d\boldsymbol{z} + \\ &\underbrace{\int p(\boldsymbol{z}|\boldsymbol{x},\boldsymbol{y}){\rm log}(\frac{p(\boldsymbol{z}|\boldsymbol{x},\boldsymbol{y})}{q(\boldsymbol{z}|\boldsymbol{x},\boldsymbol{y})})d\boldsymbol{z}}_{{\rm KL}(p(\boldsymbol{z}|\boldsymbol{x},\boldsymbol{y}) || q(\boldsymbol{z}|\boldsymbol{x},\boldsymbol{y})) \ge 0}) \\ &\ge \iint p(\boldsymbol{x},\boldsymbol{y})d\boldsymbol{x}d\boldsymbol{y} \int p(\boldsymbol{z}|\boldsymbol{x},\boldsymbol{y}){\rm log}(\frac{q(\boldsymbol{z}|\boldsymbol{x},\boldsymbol{y})}{p(\boldsymbol{z})})d\boldsymbol{z} \overset{{\rm def}}{=} \tilde{I}(\boldsymbol{z};\boldsymbol{x},\boldsymbol{y}).
\label{b-2}
\end{aligned}
\end{equation}

The lower bound is tight when $q(\boldsymbol{z}|\boldsymbol{x},\boldsymbol{y})$ converges to $p(\boldsymbol{z}|\boldsymbol{x},\boldsymbol{y})$, \textit{i.e.}, $q(\boldsymbol{z}|\boldsymbol{x},\boldsymbol{y}) = p(\boldsymbol{z}|\boldsymbol{x},\boldsymbol{y}) \Leftrightarrow p(\boldsymbol{z},\boldsymbol{x},\boldsymbol{y}) = p(\boldsymbol{x},\boldsymbol{y})q(\boldsymbol{z}|\boldsymbol{x},\boldsymbol{y})$. Then we can rewrite $\tilde{I}(\boldsymbol{z};\boldsymbol{x},\boldsymbol{y})$ as:
\begin{equation}
\begin{aligned}
    &\tilde{I}(\boldsymbol{z};\boldsymbol{x},\boldsymbol{y}) = \iint p(\boldsymbol{x},\boldsymbol{y})d\boldsymbol{x}d\boldsymbol{y} \int q(\boldsymbol{z}|\boldsymbol{x},\boldsymbol{y}){\rm log}(\frac{q(\boldsymbol{z}|\boldsymbol{x},\boldsymbol{y})}{p(\boldsymbol{z})})d\boldsymbol{z} \\ &= \iint p(\boldsymbol{x},\boldsymbol{y})d\boldsymbol{x}d\boldsymbol{y} \int q(\boldsymbol{z}|\boldsymbol{x},\boldsymbol{y}){\rm log}(\frac{q(\boldsymbol{z}|\boldsymbol{x},\boldsymbol{y})}{\iint p(\boldsymbol{z},\boldsymbol{x},\boldsymbol{y})d\boldsymbol{x}d\boldsymbol{y}})d\boldsymbol{z} \\ &= \iint p(\boldsymbol{x},\boldsymbol{y})d\boldsymbol{x}d\boldsymbol{y} \int q(\boldsymbol{z}|\boldsymbol{x},\boldsymbol{y}){\rm log}(\frac{q(\boldsymbol{z}|\boldsymbol{x},\boldsymbol{y})}{\iint p(\boldsymbol{x},\boldsymbol{y})q(\boldsymbol{z}|\boldsymbol{x},\boldsymbol{y})d\boldsymbol{x}d\boldsymbol{y}})d\boldsymbol{z} \\ &= \mathbbm{E}_{\boldsymbol{x},\boldsymbol{y}}[\int q(\boldsymbol{z}|\boldsymbol{x},\boldsymbol{y}){\rm log}(\frac{q(\boldsymbol{z}|\boldsymbol{x},\boldsymbol{y})}{\mathbbm{E}_{\boldsymbol{x},\boldsymbol{y}}[q(\boldsymbol{z}|\boldsymbol{x},\boldsymbol{y})]})d\boldsymbol{z}] \\ &= H(\mathbbm{E}_{\boldsymbol{x},\boldsymbol{y}}[q(\boldsymbol{z}|\boldsymbol{x},\boldsymbol{y})])-\mathbbm{E}_{\boldsymbol{x},\boldsymbol{y}}[H(q(\boldsymbol{z}|\boldsymbol{x},\boldsymbol{y}))].
\label{b-3}
\end{aligned}
\end{equation}

In conclusion, Theorem 2 is proven.

With the help of Theorem 2, we can obtain the variational lower bounds of Eq. (7) and Eq. (8) in the main body of the paper:
\begin{equation}
\begin{aligned}
    I(\boldsymbol{y}^{v};\boldsymbol{x}^{v},&\boldsymbol{x}^{r}) \ge H(\mathbbm{E}_{\boldsymbol{x}^{v},\boldsymbol{x}^{r}}[q(\boldsymbol{y}^{v}|\boldsymbol{x}^{v},\boldsymbol{x}^{r})]) - \\ &\mathbbm{E}_{\boldsymbol{x}^{v},\boldsymbol{x}^{r}}[H(q(\boldsymbol{y}^{v}|\boldsymbol{x}^{v},\boldsymbol{x}^{r}))] \overset{{\rm def}}{=} \tilde{I}(\boldsymbol{y}^{v};\boldsymbol{x}^{v},\boldsymbol{x}^{r}).
    \label{b-4}
\end{aligned}
\end{equation}
\begin{equation}
\begin{aligned}
    I(\boldsymbol{y}^{r};\boldsymbol{x}^{v},&\boldsymbol{x}^{r}) \ge H(\mathbbm{E}_{\boldsymbol{x}^{v},\boldsymbol{x}^{r}}[q(\boldsymbol{y}^{r}|\boldsymbol{x}^{v},\boldsymbol{x}^{r})]) - \\ &\mathbbm{E}_{\boldsymbol{x}^{v},\boldsymbol{x}^{r}}[H(q(\boldsymbol{y}^{r}|\boldsymbol{x}^{v},\boldsymbol{x}^{r}))] \overset{{\rm def}}{=} \tilde{I}(\boldsymbol{y}^{r};\boldsymbol{x}^{v},\boldsymbol{x}^{r}).
    \label{b-5}
\end{aligned}
\end{equation}

\section{Framework Details}
\subsection{Cross Prediction Alignment Learning (CPAL)}
\subsubsection{Method Description}
We devise a new bi-directional prediction alignment learning by modifying the prediction alignment learning loss proposed in OTLA-ReID \cite{OTLA-ReID}. Note that our approach doesn't employ any classifier, so the prediction here is defined as the cosine similarities between the query features and the prototypes in cross prototype memory.

\begin{figure}[!t]
	\centering
	\includegraphics[scale=0.265]{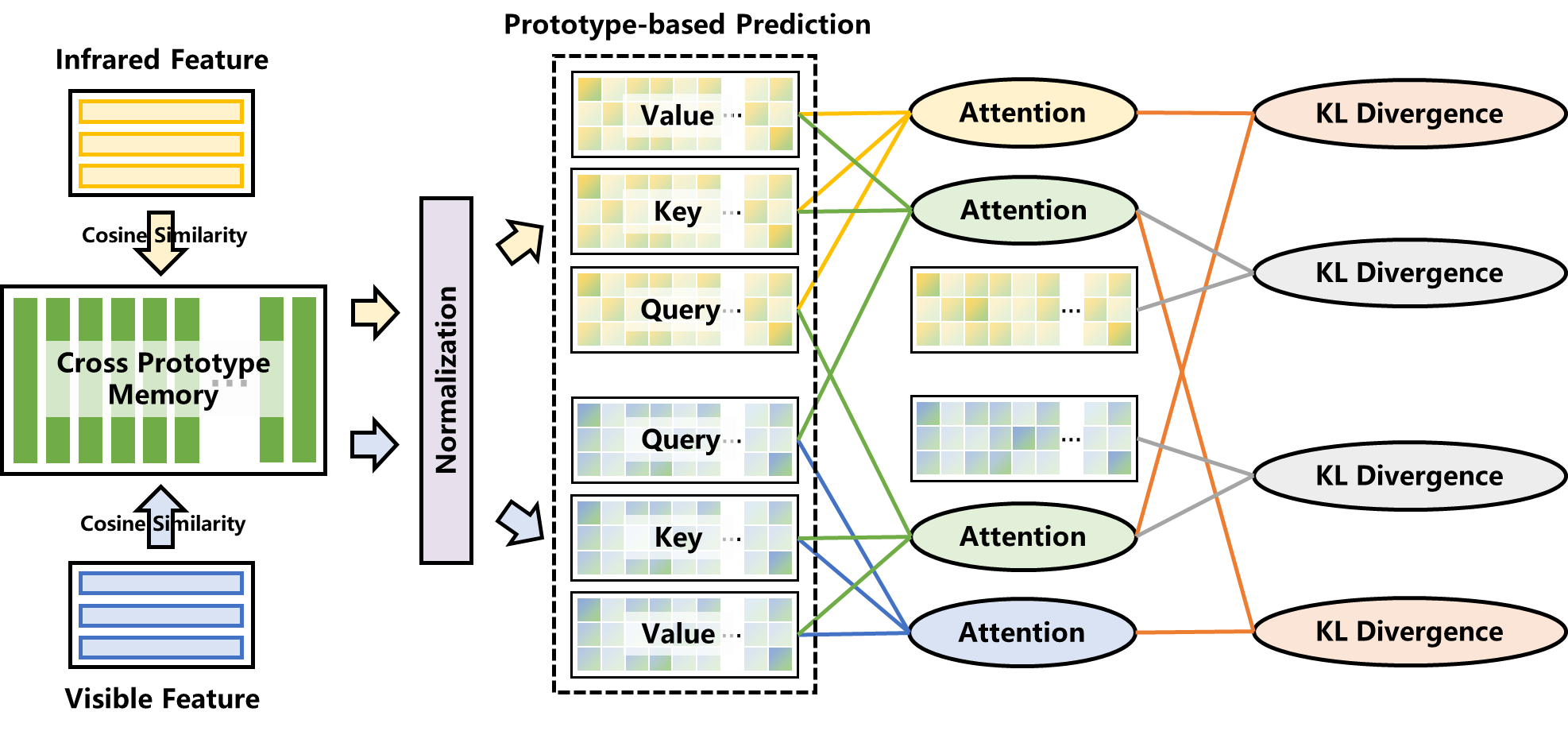}
	\caption{The diagram of the cross prediction alignment learning loss (CPAL). This loss incorporates the batch-level mix-up technique for prototype-based predictions, which can alleviate the negative effect brought by incorrectly matched cross-modality data.}
	\label{Fig:CPAL}
\end{figure}

As shown in Fig. \ref{Fig:CPAL}, for a batch of samples, we first normalize the prediction (prototype-based cosine similarities) to obtain $\boldsymbol{S}^{v} \in \mathbb{R}^{V \times N_{p}}$ and $\boldsymbol{S}^{r} \in \mathbb{R}^{R \times N_{p}}$, where the superscript $v/r$ represents visible/infrared modality and $N_p$ is the length of cross prototype memory. It is worth noting that we typically select equal-sized cross-modality samples according to the cross-modality pseudo labels generated by OTPA. Thus, visible data and infrared data have the same label distribution in each training batch (\textit{i.e.}, $R$ actually equals $V$ in our approach). 

To encourage consistent predictions on these samples, we conduct the self-attention technique by taking $\boldsymbol{S}^{v}$ as query, and $\boldsymbol{S}^{r}$ as key and value. Formally, we have:
\begin{equation}
\begin{aligned}
    \boldsymbol{S}^{vr}= {\rm Softmax}\left(\boldsymbol{S}^{v}(\boldsymbol{S}^{r})^{T} \right)\boldsymbol{S}^{r}.
\end{aligned}
\end{equation}

Indeed, we can obtain intra- and cross-modality mixed predictions in the same way, which are denoted as $\boldsymbol{S}^{rv}$, $\boldsymbol{S}^{vv}$ and $\boldsymbol{S}^{rr}$.
After that, we compute the KL divergence between the mixed predictions and the original predictions to get the alignment loss:
\begin{equation}
\begin{aligned}
    \mathcal{L}^{1}_{CPAL} = {\rm KL}(\boldsymbol{S}^{v}||\boldsymbol{S}^{vr}) + {\rm KL}(\boldsymbol{S}^{r}||\boldsymbol{S}^{rv}),
    \label{CPAL1}
\end{aligned}
\end{equation}
\begin{equation}
\begin{aligned}
    \mathcal{L}^{2}_{CPAL} = {\rm KL}(\boldsymbol{S}^{rv}||\boldsymbol{S}^{rr}) + {\rm KL}(\boldsymbol{S}^{vr}||\boldsymbol{S}^{vv}),
    \label{CPAL2}
\end{aligned}
\end{equation}
\begin{equation}
\begin{aligned}
    \mathcal{L}_{CPAL} = \mu_{1} \mathcal{L}^{1}_{CPAL} + \mu_{2} \mathcal{L}^{2}_{CPAL},
\end{aligned}
\end{equation}
where $\mu_{1}$ and $\mu_{2}$ are the coefficients (set to 0.5 and 1.0 in our experiments).

Intuitively, Eq. (\ref{CPAL1}) forces mixed cross-modality predictions to be consistent with original predictions. While Eq. (\ref{CPAL2}) forces the self-attention correlations of mixed predictions to be consistent with each other. Even though there unfortunately exists incorrectly matched data pairs, self-attention would eliminate their negative effect by promoting the instance-level alignment to batch-level alignment, so as to emphasize the truly-matched pairs while neglecting the incorrect ones. This strategy is like the mix-up technique by fusing samples from two modalities in a batch. The alignment of mixed predictions is hence encouraged to filter outliers and reduce the prediction gap between two modalities. 

\subsubsection{Visualization Analysis}
\begin{figure*}[!t]
	\centering
	\includegraphics[scale=0.54]{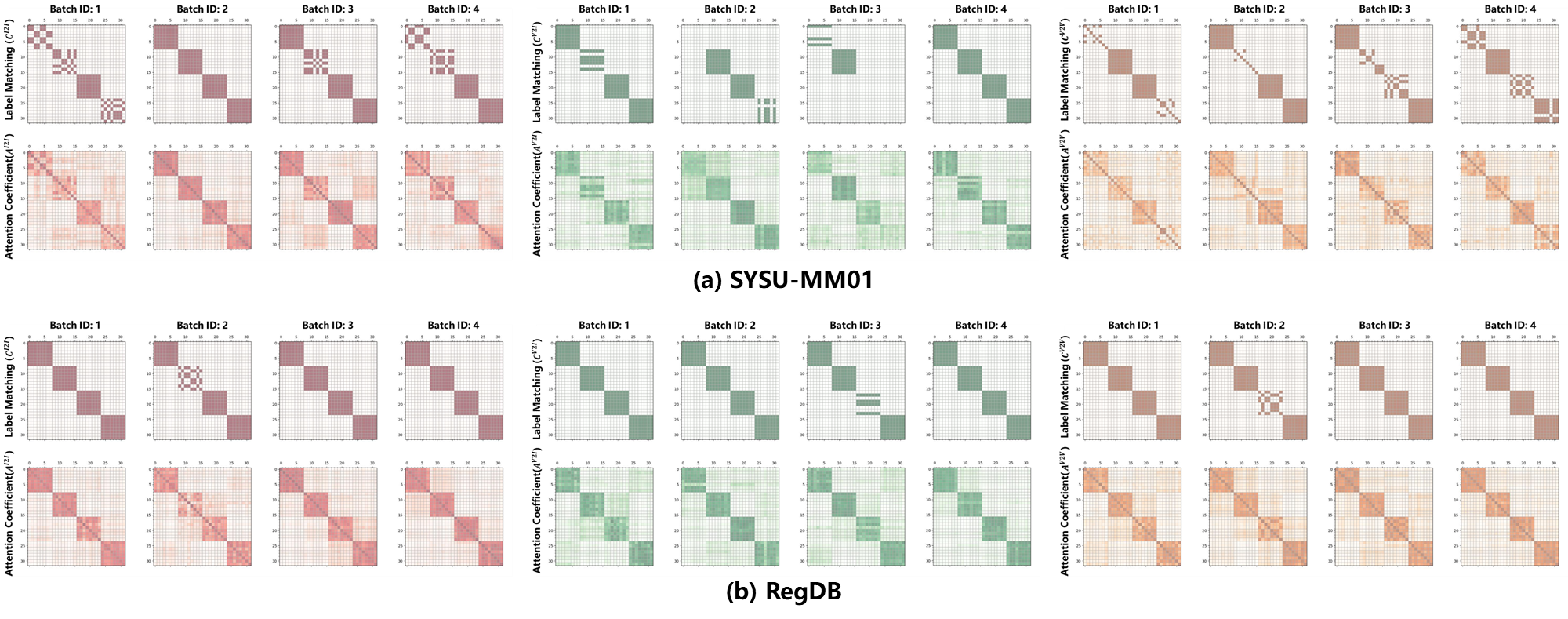}
	\caption{Visualization of cross prediction alignment learning (CPAL) on SYSU-MM01 and RegDB. The upper figures are the batch-level ground-truth binary matrices indicating whether the data pairs are matched. The bottom figures are our learned attention coefficient matrices of CPAL.}
	\label{Fig:CPALvisualization}
\end{figure*}

To better display the effectiveness of $\mathcal{L}_{CPAL}$, we visualize the attention coefficient matrices ($\boldsymbol{A}^{I2I} = {\rm softmax}(\boldsymbol{S}^r (\boldsymbol{S}^r)^T)$, $\boldsymbol{A}^{V2I} = {\rm softmax}(\boldsymbol{S}^v (\boldsymbol{S}^r)^T)$, $\boldsymbol{A}^{V2V} = {\rm softmax}(\boldsymbol{S}^v (\boldsymbol{S}^v)^T)$) shown in the bottom part of Fig. \ref{Fig:CPALvisualization}(a) and Fig. \ref{Fig:CPALvisualization}(b), and the corresponding binary ground-truth matrices ($\boldsymbol{C}^{I2I}$, $\boldsymbol{C}^{V2I}$, $\boldsymbol{C}^{V2V}$) shown in the upper part of Fig. \ref{Fig:CPALvisualization}(a) and Fig. \ref{Fig:CPALvisualization}(b). These binary matrices reflect the ground-truth correspondence of each training batch, where $1$ (fill color) indicates two samples share the same label and $0$ (not fill color) otherwise. 

From this figure, we can find that the learned attention coefficients are consistent with the binary ground-truth matrices. Even though there unfortunately exists an incorrect label, the attention matrices would eliminate its negative effect by emphasizing the truly-matched samples while neglecting the incorrect ones. Therefore, the cross prediction alignment mechanism allows us to further dig up the cross-modality correspondence. Besides, We observe a higher probability of sampling training data with matched ground-truth labels from the RegDB dataset compared to the SYSU-MM01 dataset, which further illustrates that $\mathcal{L}_{CPAL}$ is more effective on the RegDB dataset.

\begin{figure}[!t]
	\centering
	\includegraphics[scale=0.31]{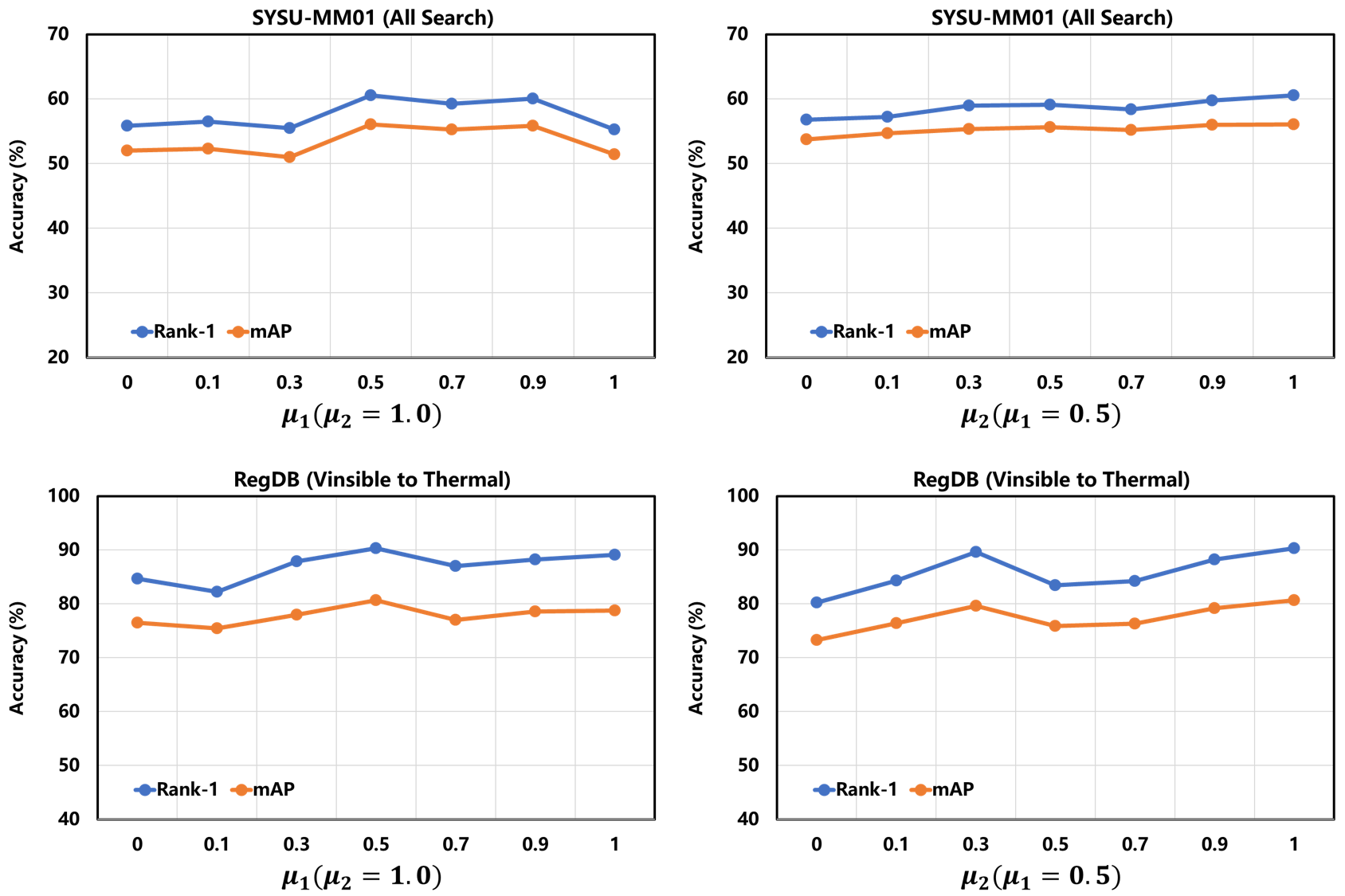}
	\caption{Sensitivity experiments corresponding to CPAL critical hyper-parameters on both SYSU-MM01 and RegDB datasets.}
	\label{Fig:sensitivityanalysis}
\end{figure}

\subsubsection{Sensitivity Analsis}
\textbf{Coefficients of $\boldsymbol{\mathcal{L}_{CPAL}}$ ($\boldsymbol{\mu_{1}}, \boldsymbol{\mu_{2}}$).} These two hyper-parameters decide the weight (importance) of the corresponding part for cross prediction alignment learning loss. We fix one of the hyper-parameters and change the other one to show its influence. Surprisingly, they are more robust than other hyper-parameters referring to Fig. \ref{Fig:sensitivityanalysis}. Besides, 0.5 for $\mu_{1}$ and 1.0 for $\mu_{2}$ seem to be proper weights for this loss.

% you can choose not to have a title for an appendix
\iffalse
% use section* for acknowledgment
\ifCLASSOPTIONcompsoc
  % The Computer Society usually uses the plural form
  \section*{Acknowledgments}
\else
  % regular IEEE prefers the singular form
  \section*{Acknowledgment}
\fi

The authors would like to thank...
\fi

% Can use something like this to put references on a page
% by themselves when using endfloat and the captionsoff option.
\ifCLASSOPTIONcaptionsoff
  \newpage
\fi

% trigger a \newpage just before the given reference
% number - used to balance the columns on the last page
% adjust value as needed - may need to be readjusted if
% the document is modified later
%\IEEEtriggeratref{8}
% The "triggered" command can be changed if desired:
%\IEEEtriggercmd{\enlargethispage{-5in}}

% references section

% can use a bibliography generated by BibTeX as a .bbl file
% BibTeX documentation can be easily obtained at:
% http://mirror.ctan.org/biblio/bibtex/contrib/doc/
% The IEEEtran BibTeX style support page is at:
% http://www.michaelshell.org/tex/ieeetran/bibtex/
%\bibliographystyle{IEEEtran}
% argument is your BibTeX string definitions and bibliography database(s)
%\bibliography{IEEEabrv,../bib/paper}
%
% <OR> manually copy in the resultant .bbl file
% set second argument of \begin to the number of references
% (used to reserve space for the reference number labels box)
% \begin{thebibliography}{1}

% \bibitem{IEEEhowto:kopka}
% H.~Kopka and P.~W. Daly, \emph{A Guide to \LaTeX}, 3rd~ed.\hskip 1em plus
%   0.5em minus 0.4em\relax Harlow, England: Addison-Wesley, 1999.

% \end{thebibliography}

\bibliographystyle{ieee_Fullname}
\bibliography{reference}

\iffalse
\begin{IEEEbiography}{Michael Shell}
Biography text here.
\end{IEEEbiography}

% if you will not have a photo at all:
\begin{IEEEbiographynophoto}{John Doe}
Biography text here.
\end{IEEEbiographynophoto}

% insert where needed to balance the two columns on the last page with
% biographies
%\newpage

\begin{IEEEbiographynophoto}{Jane Doe}
Biography text here.
\end{IEEEbiographynophoto}
\fi
% You can push biographies down or up by placing
% a \vfill before or after them. The appropriate
% use of \vfill depends on what kind of text is
% on the last page and whether or not the columns
% are being equalized.

%\vfill

% Can be used to pull up biographies so that the bottom of the last one
% is flush with the other column.
%\enlargethispage{-5in}

% that's all folks
\end{document}